\documentclass[twoside]{article}

\usepackage[accepted]{template/aistats2021}
\usepackage{microtype}
\usepackage{graphicx}
\usepackage{subfigure}
\usepackage{booktabs}
\usepackage{amssymb}
\usepackage{physics}
\usepackage{amsthm}
\usepackage{bbold}
\usepackage{hyperref}
\usepackage{paralist}

\newcommand{\rrr}{\mathbb{R}}
\newcommand{\db}{\mathcal{D}}
\newcommand{\cone}{{\mathcal{C}^{\mathcal{D}}_{\theta,v}}}
\newtheorem{Theorem}{Theorem}

\newtheorem{Lemma}{Lemma}
\newtheorem{Corollary}{Corollary}
\newtheorem{definition}{Definition}
\usepackage{extarrows}
\usepackage{xcolor}
\usepackage{graphicx}

\DeclareMathOperator*{\argmax}{arg\,max}
\DeclareMathOperator*{\argmin}{arg\,min}
\usepackage[round]{natbib}

\begin{document}
\twocolumn[
%

%

\aistatstitle{Hidden Cost of Randomized Smoothing}

\aistatsauthor{ Jeet Mohapatra$^{*}$ \And Ching-Yun Ko$^{*}$ \And Tsui-Wei Weng }

\aistatsaddress{  MIT  \And MIT \And MIT-IBM Watson AI Lab } 


\aistatsauthor{ Pin-Yu Chen \And Sijia Liu \And Luca Daniel }

\aistatsaddress{ IBM Research \And MIT-IBM Watson AI Lab \And MIT}
]

\runningauthor{Jeet Mohapatra ~ Ching-Yun Ko ~ Tsui-Wei Weng ~ Pin-Yu Chen ~ Sijia Liu ~ Luca Daniel}

\begin{abstract}
The fragility of modern machine learning models has drawn a considerable amount of attention from both academia and the public. While immense interests were in either crafting adversarial attacks as a way to measure the robustness of neural networks or devising worst-case analytical robustness verification with guarantees, few methods could enjoy both scalability and robustness guarantees at the same time. As an alternative to these attempts, randomized smoothing adopts a different prediction rule that enables statistical robustness arguments which easily scale to large networks. However, in this paper, we point out the side effects of current randomized smoothing workflows. Specifically, we articulate and prove two major points: 1) the decision boundaries of smoothed classifiers will shrink, resulting in disparity in class-wise accuracy; 2) applying noise augmentation in the training process does not necessarily resolve the shrinking issue due to the inconsistent learning objectives.




\end{abstract}

\section{INTRODUCTION}
\label{sec:intro}

\begin{table*}[h]
    \centering
    \caption{A look-up table of theoretical (\textbf{T}) and numerical (\textbf{N}) contributions in Section 4.}
    \begin{tabular}{c|cccc}
    \toprule
    region geometry & shrinking & vanishing rate $\sigma_{\text{van}}$ & shrinking rate & certified radius \\\midrule
    bounded & \textbf{T} (Thm.~\ref{thm:1}) & \textbf{T} - lower bnd. (Thm.~\ref{thm:2}) & \textbf{N} - lower bnd. (Fig.~2) & \textbf{N} - case study\\
    semi-bounded & \textbf{T} (Thm.~\ref{thm:semi1}) & not applicable & \textbf{T} - lower bnd. (Thm.\ref{thm:semi3}) & \textbf{N} - case study\\\bottomrule
    \end{tabular}
    \label{tbl:s_lookup1}
\end{table*}

Current mainstream methods to evaluate robustness of DNNs against adversarial examples~[\cite{szegedy2014intriguing,biggio2013evasion}] employ robustness verification. Such techniques can guarantee that no adversarial examples can  exist within a specified distance $r$ from a given input. As computing the largest possible $r$ has been proven to be NP-complete~[\cite{katz2017reluplex}], one popular approach is to derive a certified lower bound of $r$ through convex/linear relaxation~[\cite{hein2017formal,weng2018towards,Singh2018Fast,zhang2018crown}], which can be computed efficiently. Nevertheless, these techniques can hardly scale to state-of-the-art DNNs on ImageNet, motivating the idea of applying \textit{randomized smoothing}~[\cite{Cohen2019Certified,lecuyer2019certified,Li2019Certified,jia2019certified,lee2019tight}] (\textit{i.e.} a spatial low-pass filter) to transform the original classifier into a ``smoothed`` counterpart. This new smoothed classifier now returns the class with the highest probability by querying input data that has been purposely corrupted by isotropic Gaussian noise~$N(0,\sigma^2 \mathcal{I})$. 

Although randomized smoothing allows non-trivial robustness verification for the smoothed classifier on ImageNet, the side-effects of randomized smoothing have not yet been rigorously studied, except for a case-study of one specific binary classifier in [\cite{guch2019constructing}, p.2] and some impossibility results on accuracy-certification trade-off [\cite{yang2020randomized, blum2020random, kumar2020curse}]. 
The main motivation of this paper is to take a deep dive into the hidden cost of randomized smoothing for general multi-class classifiers.

The development of this paper is as follows: in Section~\ref{sec:related} we review basic preliminaries for adversarial robustness certification with randomized smoothing; in Section~\ref{sec:motivation} we fully expose a major hidden cost of randomized smoothing -- biased predictions, by providing evidences from both real-life and synthetic datasets; in Section~\ref{sec:main} we provide a comprehensive theory exposing the root of the biased prediction -- referred to as the \textit{shrinking phenomenon} in the remainder of the paper; in Section~\ref{sec:discussion} we hold a discussion on the effects of data augmentation on the shrinking phenomenon and implications given by our theoretical analysis. 
Table \ref{tbl:s_lookup1} summarizes our contributions.

\section{BACKGROUND}
\label{sec:related}
\subsection{Randomized Smoothing and Adversarial Robustness}
Generally, the prediction of a model for input $x_0$ is given by taking the highest output of the score function (a neural network) $g(x_0)$. 
Let $e_i$ denote the $i^{th}$ basis vector with all components 0 and the $i^{th}$ component be 1. Then the base classifier can be given as
\begin{equation}
\label{eq:classifier_base}
    f(x_0) = e_{\xi_A} ; \quad \xi_A= \argmax_j \; g_j(x_0).
\end{equation}
Correspondingly, under randomized smoothing the prediction for a model $g$ is given as the ``most likely'' standard prediction output by the model when noise is added to the input. Conventionally, the resulting classifier is referred to as the \textit{smoothed classifier} and the type of noise added to the input is denoted as the \textit{smoothing measure}. When isotropic Gaussian distribution $\mathcal{N}(0, \sigma^2 \mathcal{I})$ is used as the smoothing measure, the smoothed function $f_\sigma$ is given as 
\begin{align*}
    f_\sigma(x_0) &= e_{\xi_A} ;\\
    \quad \xi_A = \argmax_j \; \mathbb{P} \lbrack j &=  \argmax_i g_i(x) \rbrack, \, x \sim \mathcal{N}(x_0,\sigma^2\mathcal{I}). 
\end{align*}
There has been a lot of research in developing robustness verification techniques for the \textit{base classifier} in Equation~\eqref{eq:classifier_base}~[\cite{hein2017formal,weng2018towards, Gehr2018AI2,raghunathan2018certified,weng2018evaluating,wong2018provable,wang2018efficient,li2020sok}], \textit{i.e.} given $g, x_0, \xi_A$ and $p$, find the maximum value of $r$ such that $\argmax_j g_j (x_0+\delta) = \xi_A, \; \forall \|\delta \|_p \leq r$. However, due to the intrinsic hardness of the problem~[\cite{katz2017reluplex, weng2018towards,tjeng2018evaluating}], the above approaches can hardly scale to state-of-the-art deep neural networks such as ResNet-50 and VGG-19 nets. On the other hand, it is also possible to perform robustness verification on the \textit{smoothed classifier}.
To solve the problem of certification, \cite{lecuyer2019certified} first applied differential privacy techniques to derive a non-trivial lower bound of $r$ for $p = 1, 2$. The bound was later improved by~\cite{Li2019Certified} via the tools in information theory for $p = 2$. Recently,~\cite{Cohen2019Certified} proved a tighter bound of $r$ for $p = 2$ below:
\begin{align} 
\label{eq:certify_formula}
    r = \frac{\sigma}{2}\left[\Phi^{-1}(\underline{p_A})-\Phi^{-1}(\overline{p_B})\right],
\end{align}
where $\sigma$ is the smoothing factor in the Gaussian noise, $\Phi^{-1}$ is the inverse of standard Gaussian CDF, and $\underline{p_A}$ and $\overline{p_B}$ are the lower/upper bound on the probability with class $\xi_A$ and $\xi_B$ ($\xi_A$ is the top-1 class of the smoothed classifier and $\xi_B$ is the ``runner-up'' class), respectively.
In practice, \cite{Cohen2019Certified} sets $\overline{p_B}=1-\underline{p_A}$ and abstains when $\underline{p_A}<0.5$, implying that no radius can be certified in this case.

\subsection{Data Augmentation}
In the seminal work of randomized smoothing ,~\cite{Cohen2019Certified} and \cite{lecuyer2019certified} suggest to apply randomized smoothing during training (noise augmentation) for better classification accuracy. 
We first recall that a standard learning problem takes the form of $$\mathcal{R}=\mathbb{E}_{x\in\mathcal{X}}[l(f(x),h(x))],$$ where $\mathcal{X}$, $\mathcal{Y}$, $l$, $f$, and $h$ are the input space, the output space, the loss function, a neural network, and the ground-truth classifier, respectively. Given some probability distribution $\mathfrak{D}_p$ the noise smoothing risk takes the form of 
\begin{align*}
    \mathcal{R}_{\text{RS}} &= \mathbb{E}_{x\in\mathcal{X}}[l(f_\sigma(x),h(x))]\\
    &=\mathbb{E}_{x\in\mathcal{X}}[l(\mathbb{E}_{z\sim\mathfrak{D}_p}[f(x+z)],h(x))].
\end{align*}
\cite{Cohen2019Certified} motivate the use of corrupted samples during training by arguing that, when $l$ is chosen to be the cross entropy and $\mathfrak{D}_p=\mathcal{N}(0,\sigma^2I)$, the noise augmentation risk
$$\mathcal{R}_{\text{RS-train}}=\mathbb{E}_{x\in\mathcal{X}}[\mathbb{E}_{z\sim\mathcal{D}_p}[l(f_{\text{train},\sigma}(x+z),h(x))]]$$ constitutes a lower bound of $\mathcal{R}_{\text{RS}}$. We distinguish $f_{\text{train},\sigma}$ from $f$ since they are learned from different objectives. Throughout this paper, we abbreviate Gaussian noise augmentation (\textit{i.e.} $\mathfrak{D}_p$ be the Gaussian centered at the origin) as data augmentation.

\section{TWO MOTIVATING EXAMPLES}
\label{sec:motivation}
\begin{table*}[h!]
    \centering
    \caption{The mean certified radii (with $\pm$ std.) of CIFAR10 classifiers learned with data augmentation and inferred by the randomized smoothing prediction rule. ``certified radius (c)'' denotes the correct certified radius.}
    \scalebox{0.90}{
    \begin{tabular}{l|ccccccc}
    \toprule
    training $\sigma$ & 0.12 & 0.25 & 0.50 & 1.00 & 1.50 & 2.00 & 3.00\\
    min \& max  &  $(67.8\pm 1.9,$ &  $(55.4\pm 4.8,$ &  $(42.4\pm 4.8,$ &  $(20.8\pm 1.3,$ &  $(9.8\pm 1.3,$ &  $(5.4\pm 0.9,$ &  $(1.2\pm 0.8,$ \\
    class-wise acc.($\%$) & $\ \; 93.4\pm 1.3)$ &  $\ \; 89.2\pm 1.3)$ &  $\ \; 81.9\pm 2.2)$ &  $\ \; 72.8\pm 1.5)$ &  $61.2\pm 3.1)$ &  $53.2\pm 3.9)$ &  $41.0\pm 1.0)$\\\hline
    certified radius &  $0.28\pm 0.01$ &  $0.42\pm 0.02$ &  $0.51\pm 0.03$ &  $0.50\pm 0.01$ &  $0.44\pm 0.01$ &  $0.38\pm 0.01$ &  $0.32\pm 0.01$ \\
    certified radius (c) &  $0.34\pm 0.01$ &  $0.56\pm 0.01$ &  $0.80\pm 0.02$ &  $1.07\pm 0.01$ &  $1.25\pm 0.03$ &  $1.40\pm 0.03$ &  $1.80\pm 0.07$\\\bottomrule 
    \end{tabular}}
    \label{tbl:certified_radii}
\end{table*}

The major highlight of randomized smoothing techniques in the scope of adversarial robustness is its ability to provide non-trivial robustness guarantees (certified radii) for large networks. With this in mind, as pointed out in [\cite{Cohen2019Certified}, Sec. 3.2.2 last para.], for randomized smoothing with parameter $\sigma$, the maximum achievable certified radius is around $4 \sigma$, implying larger smoothing factor $\sigma$ is needed for a larger maximum achievable certified radius\footnote{One can also gain insights from that the certified radius $r$ is proportional to the smoothing factor $\sigma$ (\textit{cf. Equation~\ref{eq:certify_formula}}).}. This need is further justified in \cite{Cohen2019Certified} by pointing out the trade-off between the sample complexity and certified radii with a fixed smoothing factor. Therefore, one has to use large $\sigma$ to achieve the state-of-the-art robustness guarantees while avoiding impractical sample complexity.

In Table~\ref{tbl:certified_radii}, we validate this point by calculating the certified radii of CIFAR10 smoothed classifiers with base classifier trained with data augmentation\footnote{Throughout the paper, all the classification results and certified radii are obtained with the open-source code provided by \cite{salman2019provably}.}. In this experiment, we vary the smoothing factor $\sigma$ from 0.12 to 3.00, which is used simultaneously in data augmentation and randomized smoothing. When reporting their certified radius, we consider two metrics: 1) certified radius - the mean of all certified radii in the testing set, with the radius assigned to zero for wrongly-classified samples; and 2) correct certified radius - the mean of certified radii of correctly-classified samples in the testing set. We then see that with the increasing smoothing factor $\sigma$, the average certified radius of correctly-classified samples keeps rising from only $0.34$ to $1.80$, obtaining indeed non-trivial robustness guarantees.

On the other hand, the average certified radius of all samples climbs to around $0.5$ and then decreases to $0.32$. This is because the classification accuracy also drops as one uses larger $\sigma$, pushing more samples to have zero certified radius. In order to better understand the drop in accuracy and the affected examples, we provide a case study over a synthetic dataset.

\begin{table*}[ht]
    \caption{The class-wise accuracy ($\%$) in percentile of classifiers and smoothing factors used in \cite{Cohen2019Certified}.}
    \centering
    \scalebox{1.0}{
    \begin{tabular}{r|ccccc|ccccc}
    \toprule
    & \multicolumn{5}{c|}{CIFAR10} & \multicolumn{5}{c}{ImageNet} \\\hline
    percentile & 1st & 25th & 50th & 75th & 100th & 1st & 25th & 50th & 75th & 100th \\\hline
    $\sigma=0.00$ & 78 & 88 & 91 & 93 & 96 & 14 & 66 & 78 & 88 & 100\\
    $0.12$ & 0 & 8 & 15 & 24 & 100 & 0 & 36 & 52 & 66 & 96\\
    $0.25$ & 0 & 0 & 0 & 0 & 72 & 0 & 2 & 10 & 20 & 82\\
    $0.50$ & 0 & 0 & 0 & 0 & 98 & 0 & 0 & 0 & 0 & 56\\\bottomrule 
    \end{tabular}}
    \label{tbl:G1}
\end{table*}
\subsection{Synthetic Datasets}

Consider the binary-classification problem on the dataset ($\mathcal{X} = \mathcal{X}_1 \cup \mathcal{X}_2$) given as mixture of Gaussians:
\begin{equation*}
    \begin{aligned}
    \label{eqn:gauss_synthetic}
    & \mathcal{X}_1= (\frac{1}{2} - \epsilon) \cdot \mathcal{N}(-a, \sigma^2_o)  + \epsilon \cdot \mathcal{N}(ka, \sigma^2_o);\quad \\
    &\mathcal{X}_2= \frac{1}{2} \cdot \mathcal{N}(0, \sigma^2_o);\quad
    \end{aligned}
\end{equation*}
where $a, k, \sigma_o \in \mathbb{R}^+/\{0\}$. Then we have

\begin{Theorem}
\label{thm:decrease_rs}
Consider a classifier $f_{\text{train},\sigma_t}$ given as the naive-Bayes classifier obtained by training on the dataset $\mathcal{X}$ with data augmentation of variance $\sigma_t$. Let the class-wise accuracy of the two classes with $f_{\text{train},\sigma_t}$ using the randomized smoothing prediction rule be given as $Acc_1(\sigma_t)$ and $Acc_2(\sigma_t)$. Then we define the bias ($\Delta(\sigma_t)$) to be the gap between class-wise accuracies ($\Delta(\sigma_t) = \abs{Acc_1(\sigma_t) - Acc_2(\sigma_t)}$). For $k > \frac{1}{2\epsilon} - 1$, class I decision region grows in size at a rate of $\Theta(\sigma_t^2)$ and thus the bias is large for large $\sigma_t$. 
\end{Theorem}

It is quite well-known that using higher $\sigma$ leads to lowering of accuracy. In general, previous works have stated the existence of a robustness-accuracy trade-off. Here, we notice another interesting and quite important problem that is created by randomized smoothing: randomized smoothing based models for high values of $\sigma_t$ are biased in their predictions. Some classes are favored a lot more than others, resulting in huge difference in class-wise accuracies. 

In order to better understand the extent of the bias possible, we also study the limiting case of $\sigma_o \rightarrow 0$. This allows us to effectively study large bias without having $\sigma_t \rightarrow \infty$. In particular, we consider the dataset($\mathcal{X'}$) with probability mass function :
\begin{equation*}
\label{eqn:delta_synthetic}
    \rho(0,1)=\frac{1}{2};\quad
    \rho(-a,2)=\frac{1}{2}-\epsilon;\quad
    \rho(ka,2)=\epsilon,
\end{equation*}
with $a,k$ defined as before. For this new dataset, we see that
\begin{Theorem}
\label{thm:low_rs}
Consider a classifier $f_{\text{train},\sigma_t}$ given as the naive-Bayes classifier obtained by training on the dataset $\mathcal{X'}$ with data augmentation of variance $\sigma_t$. The bias of the classifier $f_{\text{train},\sigma_t}$ using the randomized smoothing prediction rule is $1 - \epsilon$, if $k>\frac{e^2}{\epsilon}-1$ and $\sigma_t\geq a\sqrt{\frac{k(k+1)}{2ln(2\epsilon (k+1))-\frac{2k}{k+2}}}$.
\end{Theorem}
To give intuitive understanding of the critical smoothing factor in Theorem~\ref{thm:low_rs}, we fix the scale of the dataset $a(k+1)$ to be $[0,1]$ as is common-practice in the literature~[\cite{Cohen2019Certified,salman2019provably}]. Then, we observe the shrinking effects happen at $\sigma\approx 0.7$ which is well within the realm of smoothing factors used in practice (\cite{Cohen2019Certified,salman2019provably} use smoothing factors upto $1.0$ for data augmentation and randomized smoothing).
This idea can be extended to several more general and interesting cases: a multi-class case giving accuracy $\frac{1}{c} + \epsilon$ by having class 1 with the same distribution and the rest of the classes with distributions similar to that of class 2's; and a binary-class case where adopting data augmentation does not change the optimal solution but the subsequent randomized smoothing inference still gets low accuracy for a high enough smoothing factor $\sigma$. The proofs of Theorem~\ref{thm:decrease_rs} and Theorem~\ref{thm:low_rs} are included in the supplementary materials for interested readers. 
\subsection{Real-Life Datasets}
In the existing literature, randomized smoothing remains a legitimate way of providing adversarial robustness. However, the results on the synthetic datasets suggest randomized smoothing is biased towards some classes. In order to see if the bias is present in real-life datasets we consider a new metric, namely the min and max class-wise accuracy, where we calculate separately for each class their classification accuracy and report the minimum and the maximum. In Table \ref{tbl:certified_radii} we give the performance of randomized smoothing based classifiers under the new metric. With this metric, one can then readily see that despite the increasing trend in certified radii, the class-wise accuracies becomes more imbalanced at higher smoothing factor $\sigma$. Specifically, when the smoothing factor $\sigma=0.12$, the smoothed network with base classifier being trained by data augmentation with the same magnitude of Gaussian noise classifies ``cat'' samples with 67\% accuracy and ``automobile'' samples with 92\% accuracy. However, when $\sigma=1.00$, this gap evolves to 22\% accuracy (``cat'') versus 68\% accuracy (``ship''). This comes as an unpleasant surprise since it essentially means despite the current success of randomized smoothing in adversarial robustness, the method can lead to biased predictions, causing fairness issues.

As remarked earlier, a randomized smoothing model differs from other models in two phases, data augmentation during training and smoothing during inference. As the statistical guarantees given by randomized smoothing depend on the smoothing during inference, we focus on its role in producing the bias. Before proceeding, we verify that the bias problem still persists in the absence of augmentation during training. We conduct the smoothing experiments on the pretrained models provided by Cohen et al. (2019). In Table~\ref{tbl:G1}, we report the smoothing factors $\sigma$ and corresponding class-wise accuracies (sorted ascendingly) in percentile of [1st,25th,50th,75th,100th]. 
That is, the 1st and 100th in the percentile correspond to the lowest (min) and highest (max) class-wise accuracy, respectively For CIFAR10, the [25th, 75th] percentile corresponds to the [3rd, 8th] lowest per-class accuracy. 
One can then see that originally more than $3/4$ of the classes in datasets have reasonable accuracy, which decreases as $\sigma$ goes bigger. Eventually, when $\sigma=0.5$, more than $3/4$ of the classes have $0$ accuracy. Notably, $\sigma=0.5$ is a reasonable number under the current randomized smoothing regime since the largest sigma used by~\cite{Cohen2019Certified} and \cite{salman2019provably} is $1.0$. Thus, we see that randomized smoothing produces biased results even in the absence of data augmentation during training. In the next section, we analyze how biased predictions are caused by randomized smoothing depending on the geometry of the underlying data distribution.

\section{THEORETICAL CHARACTERIZATION OF THE SHRINKING PHENOMENON}
\label{sec:main}

\begin{figure*}[h!]
    \centering
    \includegraphics[width=0.95\linewidth]{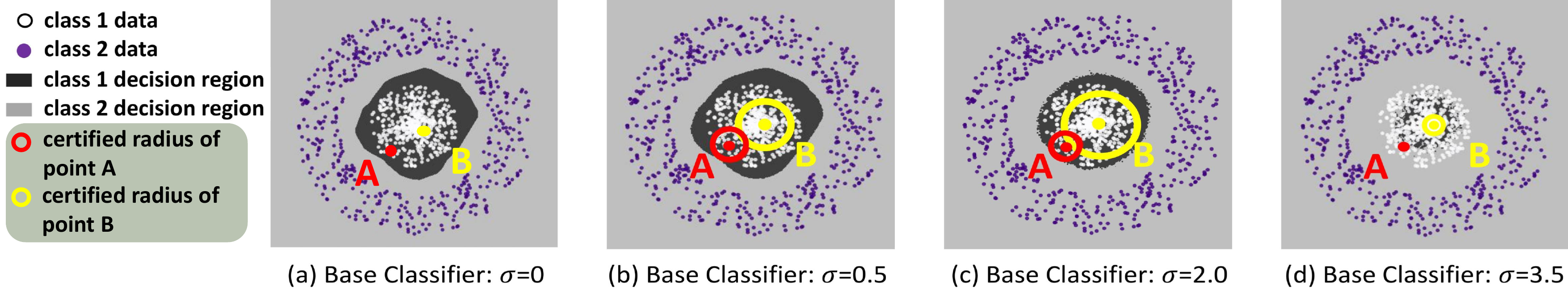}
    \includegraphics[width=0.95\linewidth]{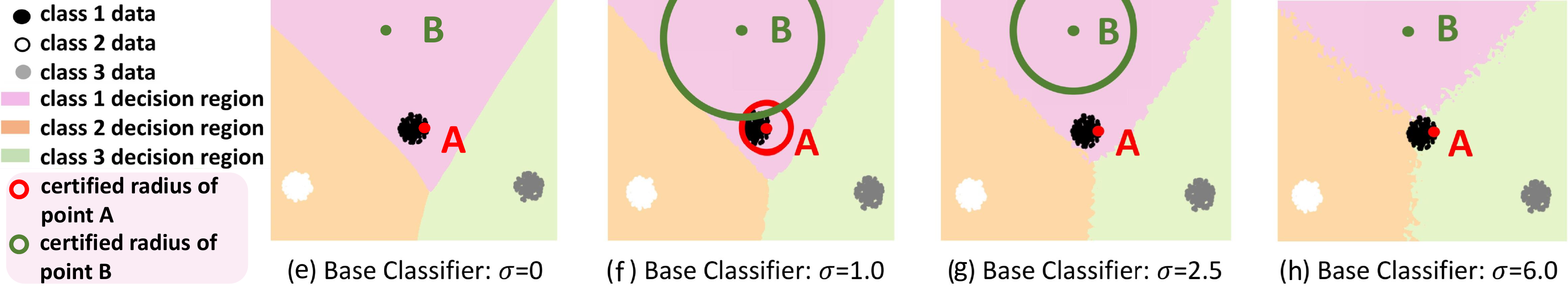}
    \caption{The 1st row shows examples of \textbf{bounded} decision regions for smoothed classifiers. The 2nd row shows examples of \textbf{semi-bounded} decision regions. The class 1 decision regions shrink as the smoothing factor $\sigma$ increases from left to right. In case (h) with larges $\sigma$, the decision region has shrunk so much that class 1 data are completely misclassified. We also plot the certified radius (Equation~\ref{eq:certify_formula}) of point $A$ and $B$ and show that it may decrease as $\sigma$ increases.}
        \label{fig:ori_circular}
\end{figure*}

Before we start our theoretical characterization, we first give a visual inspection of how randomized smoothing can change the decision regions. Specially, Figure~\ref{fig:ori_circular} illustrates two toy examples, in which the decision regions of class 1 data (the dark green region in the first row and the pink region in the second row) shrink with larger smoothing factors $\sigma$. As consequences of the shrinkage, the class-wise accuracy for class 1 data drops drastically, leading to the biased prediction.

Indeed, in this section, we aim to take a close look at this \textit{shrinking phenomenon} of randomized smoothing, uncovering the fundamental problem of the technique. Moreover, we conduct a rigorous study providing also the bounds of extreme values, beyond which the shrinking phenomenon will happen. Our results are tight and prove the prevalence of such phenomena. In order to facilitate this analysis we perform the following reductions.

\textbf{Problem Reductions.}
By the definition of randomized smoothing, the smoothed function depends on the base classifier only through the indicator function $f$. As the smoothed function $f_\sigma$ only depends on the partitioning of the input space created by the base classifier $g$, we shift our focus from the output of $g$ to how it partitions the input space, \textit{i.e.}, we are interested in characterizing all possible partitions of the input space that can lead to biased prediction as one applies randomized smoothing with a high $\sigma$. As it is hard to measure a decrease in accuracy directly from the geometry of the classifier, we approximate the decrease in accuracy using the mismatch in partitions of input space provided by $f$ and by $f_\sigma$. 

However, the problem of characterizing the partitions of the space into multiple classes is intractable. So we instead focus on tracking the behaviour of the decision boundary of a single class with respect to randomized smoothing. Without loss of generality, we set the concerned class as class 1. In this case, we analyze the misclassification rate for class 1 by the region size of the input space that is partitioned as class 1 under $f$ but not under $f_\sigma$. Considering that for any $x\in\mathbb{R}^d$, the necessary condition for it to be classified as class 1 is to have $f_\sigma(x)_1\geq \frac{1}{c}$, so we do a worst-case analysis by assuming the reformed class 1 partition is defined by exactly $f_\sigma(x)_1\geq \frac{1}{c}$.
If this overestimated reformed class 1 partition is still smaller than the original, then for sure the actual misclassification rate will be higher than the analysis herein. 

\textbf{Problem Formulation.} We formulate our problem as to characterize the ``decision regions'' that will shrink or drift after applying randomized smoothing. Formally, the decision region $\mathcal{D}$ of class 1 data is determined by the classifier $f$ via $\mathcal{D} = \{x \mid f(x)_1=1\}$. By adopting randomized smoothing, we obtain $f_\sigma(x) = \int_{x' \in \rrr^d} f(x') p(x')dx'$ with the decision region denoted by $\mathcal{D}_\sigma = \{x \mid (f_\sigma(x))_1\geq \frac{1}{c}\}$. The scope of this section is to investigate under what conditions (\textit{w.r.t.} the classifier and smoothing factor $\sigma$) will the shrinking occur.
On the whole, the shrinking effect depends highly on the geometry of the data distribution. However, considering the intractable numbers of possible decision region geometry, we will only discuss here two major classes of the geometries (\textit{bounded} in Section~\ref{subsec:bounded} and \textit{semi-bounded} in Section~\ref{subsec:semibounded}) for multidimensional data ($\textit{i.e.}$ $ d>1$). We supplement $d=1$ discussions in the supplementary materials for readers' references.
All the proofs are also deferred to the supplementary materials due to page limit.

\subsection{Bounded Decision Region}
\label{subsec:bounded}
In this section, we aim at proving the shrinking side-effects incurred by the smoothing filter when the decision region is bounded.
Formally, we say a decision region is bounded and shrinks according to the following definition:
\begin{definition}[Bounded Decision Regions]
\label{def:bounded}
If the decision region (disconnected or connected) of class 1 data is a bounded set in the Euclidean space (can be bounded by a ball of finite radius), then we call these decision regions bounded decision regions. 
\end{definition}
We denote the smallest ball that contains the original decision region of $f$ by $S_{\mathcal{D}}$ ($\mathcal{D}\subseteq S_{\mathcal{D}}$). 
Similarly, we let the smallest ball that contains the smoothed decision region (the decision region of smoothed classifier) be $S_{\mathcal{D}_\sigma}$ ($\mathcal{D}_\sigma\subseteq S_{\mathcal{D}_\sigma}$).

\begin{definition}[Shrinking of Bounded Decision Regions]
\label{def:shrinkg_bnded}
A bounded decision region is considered to have shrinked after applying smoothing filters if the radius $R_\sigma$ of $S_{\mathcal{D}_\sigma}$ is strictly smaller than the radius $R$ of $S_{\mathcal{D}}$, \textit{i.e. $R_\sigma<R$}, where $S_{\mathcal{D}}$ and $S_{\mathcal{D}_\sigma}$ are the smallest balls containing the original decision region and the smoothed decision region, respectively.
\end{definition}

 For randomized smoothing, we observe that  

\begin{Corollary}
\label{cor:1}
    The smallest ball $S_{\mathcal{D}_\sigma}$ containing the smoothed decision region is contained within the smoothed version of $S_\mathcal{D}$, \textit{i.e.} $S_{\mathcal{D}_\sigma} \subseteq (S_\mathcal{D})_\sigma$\footnote{$(S_\mathcal{D})_\sigma := \{x\mid (u_\sigma(x))_1\geq \frac{1}{c} \}$, where $u_\sigma(x) = \int_{x' \in \mathbb{R}^d} \mathbf{1}_{S_\mathcal{D}} p(x')dx'$ and $p(x')$ is the pdf of Gaussian centered at $x$.}.
\end{Corollary}

\begin{Theorem}
\label{thm:1}
A bounded decision region shrinks after applying randomized smoothing filters with large $\sigma$. Specifically, if $\sigma > \frac{R\sqrt{c}}{\sqrt{2(d-1)}} $, then $R_\sigma<R$ (\textit{cf.} Definition~\ref{def:shrinkg_bnded}).
\end{Theorem}

\textbf{Analysis of bounded decision regions with randomized smoothing.}
As we have proven that any bounded decision region shrinks after applying randomized smoothing filters, we will investigate in this part of the paper \textit{how fast} the decision region (quantified by $R_\sigma$) shrinks/vanishes. From Corollary~\ref{cor:1}, we have that the smallest ball $S_{\mathcal{D}_\sigma}$ containing the smoothed decision region is contained within the smoothed version of $S_\mathcal{D}$. Therefore we only consider the worst case when we have a ball-like decision region. 
Without loss of generality, we consider a case when the decision region of class 1 data characterized by the network function is exactly $\{x \in \rrr^d \mid \norm{x}_2 \leq R\}$. 

\begin{Theorem}[Vanishing Rate in the Ball-like Decision Region Case]
\label{thm:2}
The decision region of class 1 data vanishes when smoothing factor $\sigma_{\text{van}}>\frac{R\sqrt{c}}{\sqrt{d}}$. 
\end{Theorem}

\begin{figure}[t]
\centering
\includegraphics[width=1.0\linewidth]{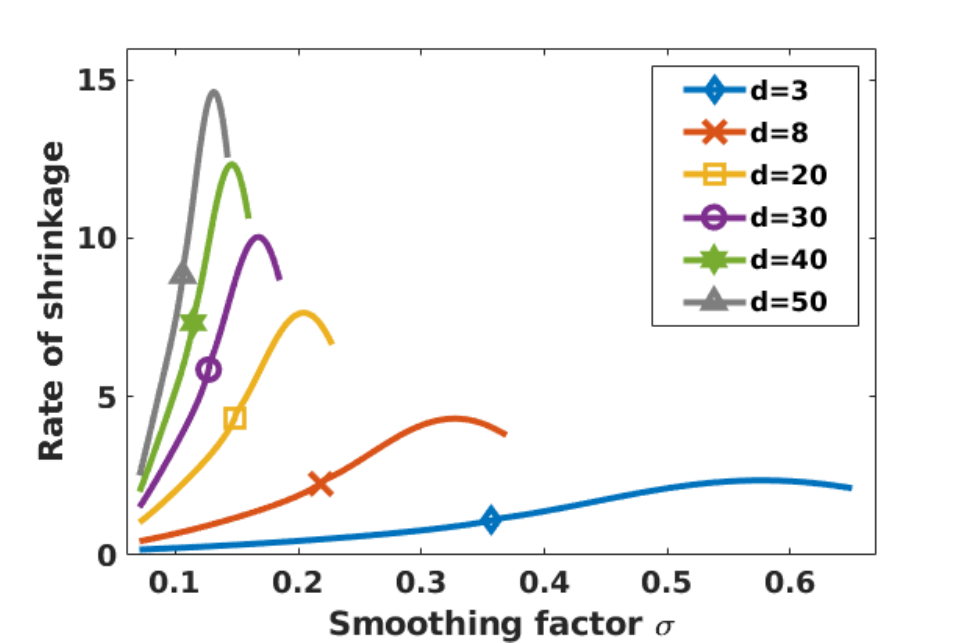}
\caption{The shrinking rate of the decision region quantified by $R_\sigma$  for different input data dimension $d$.}
\label{fig:shrinking_rate_bounded}
\end{figure}

We validate Theorem~\ref{thm:2} for binary classification ($c=2$) by substituting $R$ by $R=1$ and plot the shrinking rate (the derivatives of $R_\sigma$ with respect to $\sigma$) of the decision region as a function of the smoothing factor $\sigma$ for different input data dimensions $d=\{3,8,20,30,40,50\}$ in Figure~\ref{fig:shrinking_rate_bounded}. 
Notably, the x-axis in Figure~\ref{fig:shrinking_rate_bounded} is the varying smoothing factor $\sigma$ and the y-axis is the rate of the shrinkage  concerning class 1 decision region. 
We then see that overall the region vanishes at smaller smoothing factor $\sigma_{\text{van}}$ with the growing dimension. For example, the shrinking rate curve stops at smoothing factor $\sigma_{\text{van}}=0.651$ when $d=3$ but at smoothing factor $\sigma_{\text{van}}=0.141$ when $d=50$. We collect these vanishing smoothing factors with different data dimensions and compare with the theoretical lower bounds found in Theorem~\ref{thm:2} in the appendix to demonstrate the tightness of our theoretical lower bound. In a multi-class case, the certifiability and prediction do not follow the same setting as in~\cite{Cohen2019Certified}. For the certifiability, the effective number of classes is 2 as~\cite{Cohen2019Certified} treats it as a one vs all setting. Therefore one would be unable to certify any radius with some smoothing factor $\sigma<\sigma_{\text{van}}$ in the multi-class case. We further elaborate on this point about certifiability in Section~\ref{subsec:discussions}.

\subsection{Semi-bounded Decision Region}
\label{subsec:semibounded}

In this section, we discuss the case when the decision region is semi-bounded and is not a half-space. Formally, we say a decision region is semi-bounded and shrinks according to the following definitions:
\begin{definition}[Semi-bounded Decision Regions]
\label{def:semibounded}
If a decision region is not bounded and there exists a half-space $\mathcal{H}$ (decided by a hyperplane) that contains the unbounded decision region, then we call it semi-bounded decision region. We say a semi-bounded decision region is bounded in $v$-direction if there $\exists k\in\mathbb{R}/\infty$ such that for $\forall x\in\mathcal{D}$, $v^{T}x<k$.
\end{definition}

An illustrative example of semi-bounded decision regions is shown as Figure~\ref{fig:ori_circular}, where we have $3$ clusters of data points denoting three different classes' data and their decision regions. Observing the change in the decision region of class 1, we define ``shrinking'' as 


\begin{definition}[Shrinking of Semi-bounded Decision Regions]
\label{def:shrinking_semibnded}
A semi-bounded decision region bounded in $v$-direction is distinguished as shrinked along the direction after applying smoothing filters if the upper bound of projections of the decision region onto direction $v$ shrinks, \textit{i.e.} $\Upsilon^v_{\mathcal{D}_\sigma}<\Upsilon^v_{\mathcal{D}}$, where $\Upsilon^v_{\mathcal{D}}=\max_{x\in\mathcal{D}} v^{T}x, \Upsilon^v_{\mathcal{D}_\sigma}=\max_{x\in\mathcal{D}_\sigma} v^{T}x$.
\end{definition}

With this definition of shrinking of semi-bounded decision regions, we demonstrate in the following that any ``narrow'' semi-bounded decision region bounded in $v$-dimension will shrink along the direction (\textit{cf.} Figure~\ref{fig:ori_circular}(e-h)). We quantify the size of a decision region as follows:

\begin{definition}[$\theta,v$-Bounding Cone for a Decision Region]
A $\theta,v$ cone is defined as a right circular cone $\mathcal{C}$ with axis along $-v$ and aperture $2\theta$. Then we define the $\theta,v$-bounding cone $\mathcal{C}^\mathcal{D}_{\theta,v}$ for $\mathcal{D}$ as the $\theta,v$ cone that has the smallest projection on $v$ and contains $\mathcal{D}$, \textit{i.e.}, $\mathcal{C}^\mathcal{D}_{\theta,v} = \argmin_{\mathcal{D} \subseteq \mathcal{C}_{\theta,v}} \Upsilon^v_{\mathcal{C}_{\theta,v}}$.
\end{definition}


\begin{Theorem}
\label{thm:semi1}
A semi-bounded decision region that has a narrow bounding cone shrinks along $v$-direction after applying randomized smoothing filters with high $\sigma$, \textit{i.e.} if the region admits a bounding cone $\cone$ with $\tan(\theta)< \sqrt{\frac{(d-1)}{2c\log(c-1)}}$, then for $\sigma > (\Upsilon^v_\cone - \Upsilon^v_\db)\tan(\theta)\sqrt{\frac{c}{d-1}}\cdot\frac{2(d-1)}{(d-1) - 2\tan^2(\theta)c\log(c-1)}$, $\Upsilon^v_{\db_\sigma}<\Upsilon^v_{\db}$  (\textit{cf.} Definition~\ref{def:shrinking_semibnded}).
\end{Theorem}
Concretely, the narrowness condition (the larger the easier to fulfil) of the cone for MNIST dataset~[\cite{LeCun1998The}] relaxes to $0.43\pi = 76.7^\circ$ , meaning that if any single class's decision region can be bounded by a $\theta,v$ cone with $\theta$ being less than $76.7^\circ$, then shrinking effect happens. Correspondingly, this narrowness condition  for CIFAR10 dataset~[\cite{LeCun1998The}] is $0.46\pi=83.2^\circ$ and $0.42\pi=75.2^\circ$ for ImageNet dataset~[\cite{russakovsky2015imagenet}]. Notably, for binary classification tasks ($c=2$), according to Theorem~\ref{thm:semi1}, the condition for shrinking reduces to $\tan(\theta)<\infty$ that implies $\theta< \pi/2$. In other words, when there are only two classes, as long as the semi-decision region is not a half-space, it \textbf{will} shrink.

\textbf{Analysis of the semi-bounded case with randomized smoothing.}
As in Section~\ref{subsec:bounded}, we conduct the analysis using the worst-case ball-like bounded decision region, here we correspondingly consider a solid right circular cone along the $v$ direction. 
The shrinkage in this case serves as a non-trivial lower bound.
Without loss of generality, we consider a $\theta,v$ solid right circular cone $\{x\in\mathbb{R}^d\mid v^Tx-\|v\|\|x\|cos(\theta)\leq 0\}$ as the decision region $\mathcal{D}$ of class 1 data, where $-v=[0,\ldots,0,1]^T\in\mathbb{R}^d$.
Since the semi-bounded decision region is unbounded and will shrink but will not vanish, we emphasize in this section only on giving the shrinking rate with respect to the smoothing factor $\sigma$, the number of classes $c$, the angle $\theta$, and the data dimension $d$ with randomized smoothing. Two major theorems regarding the shrinking rate in the solid cone-like decision region are:
\begin{Theorem}
\label{thm:semi2}
The shrinkage of class 1 decision region is proportional to the smoothing factor, \textit{i.e.} \mbox{$\Upsilon^v_{\mathcal{D}}-\Upsilon^v_{\mathcal{D}_{\sigma}}\propto\sigma$}. 
\end{Theorem}
With the above Theorem~\ref{thm:semi2}, we can fix the smoothing factor to $\sigma=1$ and further obtain a lower bound of the shrinking rate \textit{w.r.t} $c$, $\theta$, and $d$:
\begin{Theorem}
\label{thm:semi3}
The shrinking rate of class 1 decision region is at least $\sqrt{\frac{d-1}{c\tan^2(\theta)}}\cdot\frac{(d-1) - 2\tan^2(\theta)c\log(c-1)}{2(d-1)}$, \textit{i.e.} $\frac{\Upsilon^v_{\db_\sigma}-\Upsilon^v_{\db_{\sigma+\delta}}}{\delta}>\sqrt{\frac{d-1}{c\tan^2(\theta)}}\cdot\frac{(d-1) - 2\tan^2(\theta)c\log(c-1)}{2(d-1)}$.
\end{Theorem}


\subsection{Remarks on Certified Radii}
\label{subsec:discussions}
\begin{figure*}[t]
\centering
\subfigure[]{
\begin{minipage}{0.48\linewidth}
\centering
\includegraphics[width=0.8\linewidth]{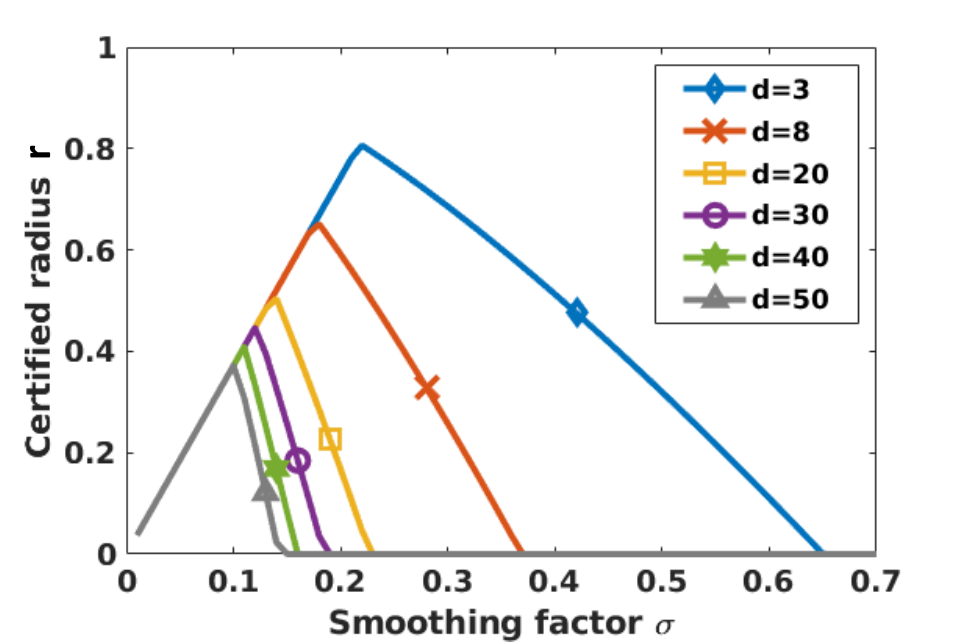}
\end{minipage}\label{fig:radius_dimension}}
\subfigure[]{
\begin{minipage}{0.48\linewidth}
\centering
\includegraphics[width=0.8\linewidth]{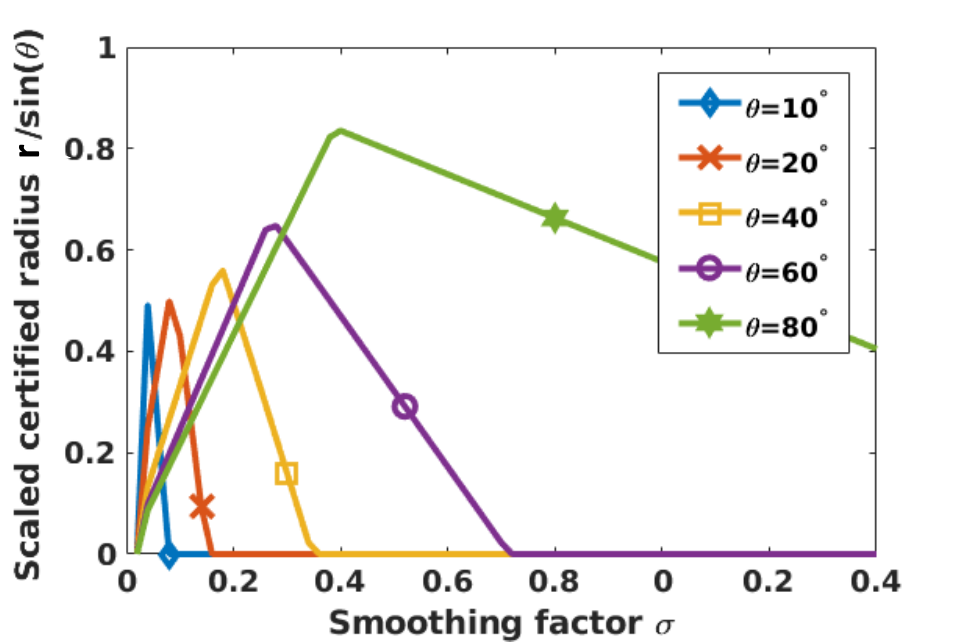}    
\end{minipage}\label{fig:radius_theta}}
\caption{(a) The certified radius $r$ of the point at the origin for different input data dimension $d$; (b) The scaled certified radius $\frac{r}{\sin(\theta)}$ of a point on the axis $v$ for cones with different apertures ($2\theta$).}
\end{figure*}


In the case of bounded decision region, the point at the origin has the highest probability to be classified as class 1 (see supplementary materials for the proof). Therefore when it has less than $0.5$ probability to be classified as class 1, the decision region vanishes and no point can be certified (certified radius $r=0$).
Specifically,
Figure~\ref{fig:radius_dimension} describes the certified radius $r$ of the point at the origin using Equations~\eqref{eq:certify_formula} as a function of the smoothing factor $\sigma$ and shows that the maximum certified radius (the peak) decreases with the increasing dimension. 
We include complete certified radius behavioral plots for different dimensions in the supplementary materials. As training samples are normally scaled in practice, they lie within a ball of radius $R\leq \sqrt{d}/2$. According to Theorem~\ref{thm:2}, for this ball, the upper bound of $\sigma_{\text{van}}$ is $1/\sqrt{2} \approx 0.707$. So in practice, if the decision region of any class lies within the volume spanned by the training samples, its certifiable region vanishes for $\sigma \geq 0.708$, regardless of the input-space dimension $d$.

In the case of semi-bounded decision region, the point on the axis has the highest probability to be classified as class 1, thus we study the certified radius of a point  $x_0=[0,\ldots,0,1]$ as a function of cone narrowness $\theta$ and smoothing factor $\sigma$. Acknowledging that the minimum distance from $x_0$ to $\theta,v$ cones is $\sin(\theta)$, we show in Figure~\ref{fig:radius_theta} the scaled certified radius $r/\sin(\theta)$ when $d=25$. One can then readily verify that overall the peak scaled certified radius decreases with $\theta$, \textit{e.g.} the scaled certified radius at $x_0$ can be as large as $0.84$ when $\theta=80^\circ$, while it is at most $0.49$ when $\theta=10^\circ$.
Moreover, we point out that certified radii drop to zero when we keep increasing the smoothing factor $\sigma$ -  the ``narrower'' (smaller $\theta$) the decision region is, the faster they drop to zero. We discuss the effect of input data dimension $d$ on the certified radius in the supplementary materials.

\begin{table*}[ht!]
    \caption{The minimum and maximum class-wise accuracy ($\%$) of CIFAR10 classifiers learned with data augmentation and inferred by the randomized smoothing prediction rule. The smaller the gap between the maximum and the minimum class-wise accuracies is, the better.}
    \centering
    \scalebox{1.0}{
    \begin{tabular}{r|cc|cc|cc|cc}
    \toprule
    \#Augmentation Points & \multicolumn{2}{c|}{1 (standard)} & \multicolumn{2}{c|}{10} & \multicolumn{2}{c|}{25} & \multicolumn{2}{c}{50}\\\hline
    class-wise acc. & min & max & min & max & min & max & min & max\\\hline
    $\sigma=0.12$ & 67 & 94 & 76 & 96 & 78 & 96 & 68 & 97 \\
    $0.25$ & 55 & 90 & 68 & 92 & 65 & 93 & 48 & 84 \\
    $0.50$ & 46 & 84 & 51 & 84 & 52 & 81 & 0 & 87 \\
    $1.00$ & 22 & 73 & 28 & 74 & 27 & 72 & 3 & 64 \\\bottomrule 
    \end{tabular}}
    \label{tbl:heavy augmentation}
\end{table*}

Interestingly, the certified radii increase with the growing smoothing factor $\sigma$ but begin to decrease at certain point - larger certified radius can normally be obtained by larger smoothing factor $\sigma$ according to Equation~\eqref{eq:certify_formula} but the dominance is taken over by the vanishing decision region when the $\sigma$ is enough-close to $\sigma_{\text{van}}$. This also explains the eventual decrease in the average certified radius seen in Table \ref{tbl:certified_radii}. For small values of $\sigma$ the average certified radius keeps increasing to a point ($\sigma_{thres} \in [0.50, 1.00]$) after which the effect of the vanishing decision region reduces the average certified radius.

\section{EFFICACY OF DATA AUGMENTATION}
\label{sec:discussion}

As Section~\ref{sec:main} proves that the biased prediction comes from the shrinking phenomenon of randomized smoothing, we want to hold a discussion herein investigating whether the state-of-the-art workflow for boosting randomized smoothing accuracy can solve this issue. 

\subsection{Counteracting Shrinking Effect Of Smoothing}
Through the above arguments, we see that to counter-effect the shrinkage induced by randomized smoothing, one will want to obtain larger decision regions for geometrically compact classes. Assuming a well-balanced distribution of classes, compact classes have a larger number of points near the margin compared to more spread-out classes. As a result, data augmentation expands the compact classes a lot more compared to other classes, partially alleviating the shrinking issue caused by smoothing. As a result, we see that the experiments in Table \ref{tbl:G1} (without data augmentation) have a much bigger bias in prediction compared to the experiments in Table \ref{tbl:heavy augmentation} column ``1-standard'', \textit{e.g.} when $\sigma=0.12$, Table \ref{tbl:G1} reads 0 versus 100 and Table \ref{tbl:heavy augmentation} reads 67 versus 94. 

However, it is important to note that the two effects do not exactly cancel each other out. Especially for high values of $\sigma$, the expansion caused by data augmentation can cause some of the more compact classes to dominate over all other classes, resulting in a highly biased classifier. Table \ref{tbl:heavy augmentation} shows that the bias of the classifier consistently increases with increasing values of $\sigma$ regardless of the number of augmenting points used. This signals two important observations: the need to limit the use of high values of smoothing factor $\sigma$ and the need for a data geometry dependent augmentation scheme to properly counteract the shrinking effect caused by smoothing.

\subsection{Heavy Data Augmentation}
Besides showing the minimum and maximum class-wise accuracies of multiple CIFAR10 classifiers trained with standard data augmentation, we also give in Table~\ref{tbl:heavy augmentation} the corresponding accuracies for an enhanced version of data augmentation. Essentially, different from the standard data augmentation implementation, where only one point is used to estimate the expectation $\mathbb{E}_{z\sim\mathcal{D}_p}[l(f_{\text{train},\sigma}(x+z),h(x))]$ inside $\mathcal{R}_{\text{RS-train}}$, we evaluate the expectation using \{10, 25, 50\} points, reducing the estimation bias. We denote this scheme as heavy data augmentation. Using a larger number of augmentation allows us to approximate the augmented distribution more closely and remove any unnecessary bias that is caused by using a bad approximation of the data augmentation. The results in Table \ref{tbl:heavy augmentation} show that the bias is slightly reduced by using a larger number \{10, 25\} of augmentation points but the problem still remains. Particularly, we see the relative improvement from increasing augmentation points becomes smaller with a larger smoothing factor $\sigma$. It is also worth noting that the gap in accuracies blows when we use up to 50 heavy data augmentation points, performing even worse than using the standard data augmentation. These observations signal it to be a more fundamental problem relating to the way we do data augmentation.


\section{CONCLUSION} 
\label{sec:conclusion}
In this paper, we provide a theoretical characterization showing that randomized smoothing during inference can lead to a drastic gap among class-wise accuracies, even when it is included in the training phase. In addition, we observe that the smoothing during inference is very sensitive to the distribution of the data and can have wildly-different effects on different classes depending on the data geometry. A similar analysis could be extended to other smoothing functions in addition to Gaussian smoothing. Crucially, our results point out the need for limiting the use of large values of $\sigma$, as well as the need for data-geometry dependent noise augmentation schemes.



\section*{Acknowledgements}
J. Mohapatra, C.-Y Ko, and L. Daniel are supported by MIT-IBM Watson AI Lab.  


\bibliography{RS.bib}

\renewcommand{\thetable}{S\arabic{table}}  
\renewcommand{\thefigure}{S\arabic{figure}} 

\onecolumn
\aistatstitle{Supplementary Materials}
\appendix
\tableofcontents

\newpage

\section{Proofs for ``TWO MOTIVATING EXAMPLES''}
\subsection{Proofs for Sec 3.1}

\begin{Lemma}
\label{s_lemma:cdfineq}
$\Phi[x]+\Phi[\frac{1}{x}]\geq1.5$ with equality holds iff $x\in\{0,\infty\}$.
\begin{proof}
    Let $f(x) = \Phi[x]+\Phi[\frac{1}{x}]$. We observe that $f(x) = f(1/x)$ by definition. So, it is sufficient to show that for $x$ in the interval $(1, \infty)$, $f(x) \geq 1.5$ with equality at $x \rightarrow \infty$. We prove this by showing that in the interval $(1, \infty), f(x)$ is strictly decreasing and $\lim_{x \rightarrow \infty} f(x) = \lim_{x \rightarrow \infty} \Phi(x) + \Phi(1/x) = \Phi(\infty) + \Phi(0) = 1 + 0.5 = 1.5$. To show f(x) is strictly decreasing we proceed by taking the derivative wrt x,
    $$\frac{d}{dx} f(x) = \frac{e^{-\frac{x^2}{2}}}{\sqrt{2\pi}} - \frac{e^{-\frac{1}{2x^2}}}{x^2 \sqrt{2\pi}} $$
    we show that for the interval $(1, \infty)$ this derivative is less than 0. So, we need to show that 
    \begin{align*}
        & \frac{e^{-\frac{x^2}{2}}}{\sqrt{2\pi}} - \frac{e^{-\frac{1}{2x^2}}}{x^2 \sqrt{2\pi}} < 0 \\
        \Leftrightarrow\quad & x^2 e^{-\frac{x^2}{2}} <e^{-\frac{1}{2x^2}}\\
        \Leftrightarrow\quad & \log(x^2) + \frac{1}{2x^2} < x^2 \\
        \text{Let }  &t = \log(x^2), x > 1 \rightarrow t > 0 \\
        \Leftrightarrow\quad & 2t < e^{t} - e^{-t}
    \end{align*}
    \noindent This holds for $t > 0$ as we have that at $t = 0$. $2\cdot 0 = 0 = e^{0} - e^{-0} $ and $2t$ increases at a rate of 2 while $e^{t} - e^{-t}$ increases at a rate of $e^{t} + e^{-t} > 2\cdot \sqrt{e^{t}\cdot e^{-t}} = 2$ as $t>1 \rightarrow e^{t} \neq e^{-t}$. Finally for $x = 1$, we calculate $f(x) \approx 1.6829 > 1.5$.
\end{proof}
\end{Lemma}

\begin{Theorem}
\label{thm:decrease_rs}
Consider a classifier $f_{\text{train},\sigma_t}$ given as the naive-Bayes classifier obtained by training on the dataset $\mathcal{X}$ with data augmentation of variance $\sigma_t$. Let the class-wise accuracy of $f_{\text{train},\sigma_t}$ using the randomized smoothing prediction rule be given as $Acc_1(\sigma_t), Acc_2(\sigma_t)$. Then we define the bias ($\Delta(\sigma_t)$) to be the gap between class-wise accuracies ($\Delta(\sigma_t) = \abs{Acc_1(\sigma_t) - Acc_2(\sigma_t)}$). For $k > \frac{1}{2\epsilon} - 1$, class I decision region grows in size at a rate of $O(\sigma_t^2)$ and thus the bias is large for large $\sigma_t$. 
\begin{proof}
    In order to determine the accuracies we start by looking at the decision regions given by the two classifiers. We show that the decision region of class 1 increases with increasing $\sigma$ effectively increasing the bias by increasing the class 1 accuracy while decreasing the class 2 accuracy.
    
    From the structure of the dataset it is easy to show that the naive Bayes classifier yield decision regions:\\
    class 1 : $[-(\frac{a}{2} + c_0(\sigma)), \frac{ka}{2} + d_0(\sigma)]$ \\
    class 2 : $[-\infty , -(\frac{a}{2} + c_0(\sigma))] \cup [ \frac{ka}{2} + d_0(\sigma), + \infty]$\\
    The likelihood ratio function $r_\sigma(x) = \frac{p(x \in \text{ class }2)}{p(x\in \text{ class }1)} = (1 - 2\epsilon) e^{\frac{-a(2x+a)}{2\sigma^2}} + 2\epsilon e^{\frac{(2x-ka) ka}{2\sigma^2}}$. This is a convex function is $x$ resulting in the previous form of decision regions.   Thus, we get the following decision regions after smoothing, class 1 $[-(\frac{a}{2} + c_1(\sigma)), \frac{ka}{2} + d_1(\sigma)]$ and the rest being class 2. 
    
    In this case we show that for $c_0(\sigma)$ grows at $\Theta(\sigma^2)$ with increasing $\sigma$ by establishing a lower bound and upper bound which both grow at the rate of $O(\sigma^2)$. \\
    For the lower bound consider the function $r^u_\sigma(x) = (1 - 2\epsilon) e^{\frac{ax}{\sigma^2}} + 2\epsilon e^{\frac{-kax}{\sigma^2}} > r_\sigma(-(\frac{a}{2} + x) )$. If for any $c_l(\sigma)$ we have $r^u_\sigma(c_l(\sigma)) = 1$, then $r_\sigma(-(\frac{a}{2} + c_l(\sigma))) < 1$. Thus, we see that using the convexity argument from before $c_0(\sigma) > c_l(\sigma)$. But it is easy to see that if $c_l(1)$ is a solution of the equation $r^u_1(x) = 1$ at $\sigma = 1$, then $\sigma^2 c_l(1)$ is a solution for $r^u_\sigma (x) = 1$. \\
    As $r^u_1$ is a continuous function with $r^u_1(0) = 1$ and $\lim_{x \rightarrow \infty} r^u_1(x) \rightarrow \infty$, it is sufficient to show that $\frac{d}{dx}r^u_1(0) = a(1 - 2\epsilon(k+1)) < 0$ (follows from the case condition) to show that $r^u_1(x) = 1$ has a positive real solution and consequently $c_0(\sigma) > \sigma^2 c_l(1) = O(\sigma^2)$. 
    From the likelihood function, we can also clearly see that $r_\sigma(-( \frac{a}{2} + x)) > (1-2\epsilon)e^{\frac{ax}{\sigma^2}}$. Using this we can establish that $c_0(\sigma) < \frac{\sigma^2-\log(1-2\epsilon)}{a}$ making $c_0(\sigma) = \Theta(\sigma^2)$.
    
    As $d_0(\sigma) \geq 0$, we have that for all $\sigma \in (0, \infty)$ the size of the interval $[-(\frac{a}{2} + c_0(\sigma)), \frac{ka}{2} + d_0(\sigma)]$ is bigger that $C\sigma^2 + C$ for some positive constant $C$. Thus, we have that at $x = -(\frac{a}{2} + c_0(\sigma) - \frac{1}{C})$ the probability $x \in $ Class I after smoothing is given as $\Phi(\frac{C\sigma^2}{\sigma_t}) - \Phi(\frac{-1/C}{\sigma_t})$. By Lemma \ref{s_lemma:cdfineq}, we get that $\Phi(\frac{C\sigma^2}{\sigma_t}) - \Phi(\frac{-1}{C\sigma_t}) > \Phi(\sigma_t C) - \Phi(\frac{-1}{C\sigma_t}) = \Phi(\sigma_t C) - (1 - \Phi(\frac{1}{C\sigma_t})) = \Phi(\sigma_t C) + \Phi(\frac{1}{C\sigma_t}) - 1 > 0.5 $. Thus, we have $c_1(\sigma) > c_0(\sigma) - \frac{1}{C}$. Combining this with the fact that clearly $c_0(\sigma) > c_1(\sigma)$, we have $c_1(\sigma) \in (c_0(\sigma) - \frac{1}{C}, c_0(\sigma))$ and similarly, we also have $d_1(\sigma) \in (d_0(\sigma) - \frac{1}{C}, d_0(\sigma))$. This also gives us $c_1(\sigma) = \Theta(\sigma^2) = \Theta(\sigma_t^2).$
    
    Consider the function $f_x(\sigma) = r_\sigma(x)$. By differentiating this function wrt $\sigma$ we see that it has only one extremum point. Using the fact that $\lim_{\sigma \rightarrow \infty} f_x(\sigma) = 1$ we have that if for any $x$, $f_x(\sigma) = 1$ then  we see that there the extremum point lies between $\sigma$ and $\infty$. If for any $\sigma' > \sigma$, $f_x(\sigma') = 1$, then there would be a two extremum points one between $\sigma, \sigma'$ and another between $\sigma', \infty$. Using this along with the continuity of $f_x$ we get that either $f_x(\sigma') < 1 \forall \sigma' > \sigma$ or $f_x(\sigma') > 1 \forall \sigma' > \sigma$. We can further use the fact that $f_x(0) \rightarrow \infty$ to see that $f_x$ is decreasing at $\sigma$ making $f_x(\sigma') < 1 \forall \sigma' > \sigma$. Thus, we see that $d_0(\sigma), c_0(\sigma)$ are increasing functions of $\sigma$. Combining this with the previous result shows that the decision region of class I after smoothing increases at $O(\sigma_t^2)$.
    
    For the bias we see that as $\sigma_t \rightarrow \infty$, class I at least occupies the region $(-\infty, \frac{ka}{2}]$ while class II occupies at most the region $(\frac{ka}{2}, \infty)$. As a result the bias is lower bounded by $(1 - \Phi(\frac{-ka}{2\sigma_o})) - \epsilon(1 - \Phi(\frac{-ka}{2\sigma_o})) = (1 - \epsilon)(1 - \Phi(\frac{-ka}{2\sigma_o})) $ which is very high.
\end{proof} 
\end{Theorem}

\begin{Theorem}
\label{thm:low_rs}
Consider a classifier $f_{\text{train},\sigma_t}$ given as the naive-Bayes classifer obtained by training on the dataset $\mathcal{X'}$ with data augmentation of variance $\sigma_t$. The bias of the classifier $f_{\text{train},\sigma_t}$ using the randomized smoothing prediction rule is $1 - \epsilon$, if $k>\frac{e^2}{\epsilon}-1$ and $\sigma_t\geq a\sqrt{\frac{k(k+1)}{2ln(2\epsilon (k+1))-\frac{2k}{k+2}}}$.
\begin{proof}
At $x=-a$, we see that if the decision region for class 1 is $[-(a+c),\frac{ka}{2}+d]$, then the probability after smoothing is 
\begin{align*}
    g(-a,1)&=\int_{x'\in\mathbb{R}^d} d(-a,x')\psi(x',1)dx'\\
    &=\int_{-(a+c)}^{\frac{ka}{2}+d}d(-a,x')dx'\\
    &=\int_{-\infty}^{\frac{ka}{2}+d}d(-a,x')dx'-\int_{-\infty}^{-(a+c)}d(-a,x')dx'\\
    &=\Phi(\frac{\frac{ka}{2}+d+a}{\sigma})-\Phi(\frac{-c}{\sigma})\\
    &\geq \Phi(\frac{\frac{k+2}{2}a}{\sigma})-\Phi(\frac{-c}{\sigma}) \quad (\text{if~} d\geq 0)\\
    &\geq \Phi(\frac{\frac{k+2}{2}a}{\sigma})-\Phi(-\frac{\sigma}{\frac{k+2}{2}a}) \quad (\text{if~} c\geq \frac{2\sigma^2}{(k+2)a})\\
    &>0.5. \quad (\text{by Lemma~\ref{s_lemma:cdfineq}})
\end{align*}
That's said, the bias will be atleast $1-\epsilon$ if $d\geq 0$ and $c\geq \frac{2\sigma^2}{(k+2)a}$ are true. We now check for $d\geq 0$: for $x\in[0,\frac{ka}{2}]$,
\begin{equation*}
\begin{aligned}
    \psi(x,1)&=\int_{x'\in\mathbb{R}^d} d(x,x')\rho(x',1)dx'&&=d(x,0)\rho(0,1) \\ &=\frac{1}{\sqrt{2\pi\sigma^2}}[\frac{1}{2}e^{-\frac{x^2}{2\sigma^2}}] &&=\frac{1}{\sqrt{2\pi\sigma^2}}[(\frac{1}{2}-\epsilon)e^{-\frac{x^2}{2\sigma^2}}+\epsilon e^{-\frac{x^2}{2\sigma^2}}]\\
    &>\frac{1}{\sqrt{2\pi\sigma^2}}[(\frac{1}{2}-\epsilon)e^{-\frac{(x+a)^2}{2\sigma^2}}+\epsilon e^{-\frac{(ka-x)^2}{2\sigma^2}}]    &&=d(x,-a)\rho(-a,2)+d(x,ka)\rho(ka,2) \\ &=\int_{x'\in\mathbb{R}^d} d(x,x')\rho(x',2)dx'=\psi(x,2),
\end{aligned}
\end{equation*}

implying $x\in[0,\frac{ka}{2}]$ belongs to class 1 for the naive bayes classifier. Therefore the decision region for class 1 extends at least to $\frac{ka}{2}+d$ with $d\geq 0$. Next, we check for $c\geq \frac{2\sigma^2}{(k+2)a}$: at $x=-a-\frac{2\sigma^2}{(k+2)a}$, the probability is 
\begin{equation*}
\begin{aligned}
    \psi(-a-\frac{2\sigma^2}{(k+2)a},1)&=\int_{x'\in\mathbb{R}^d} d(-\frac{2\sigma+a}{(k+2)\frac{a}{\sigma}},x')\rho(x',1)dx'
    &&=\frac{1}{\sqrt{2\pi\sigma^2}}[\frac{1}{2}e^{-\frac{x^2}{2\sigma^2}}]|_{x=-a-\frac{2\sigma^2}{(k+2)a}}\\
    \psi(-a-\frac{2\sigma^2}{(k+2)a},2)&=\int_{x'\in\mathbb{R}^d} d(-\frac{2\sigma+a}{(k+2)\frac{a}{\sigma}},x')\rho(x',2)dx'
    &&=\frac{1}{\sqrt{2\pi\sigma^2}}[(\frac{1}{2}-\epsilon)e^{-\frac{(x+a)^2}{2\sigma^2}}+\epsilon e^{-\frac{(ka-x)^2}{2\sigma^2}}]|_{x=-a-\frac{2\sigma^2}{(k+2)a}}.
\end{aligned}
\end{equation*}

Therefore we see that $\psi(-a-\frac{2\sigma^2}{(k+2)a},1)>\psi(-a-\frac{2\sigma^2}{(k+2)a},2)$ if
\begin{equation*}
\begin{aligned}
    & (1-2\epsilon)e^{(\frac{a}{\sigma})^2\frac{1}{2}+\frac{2}{k+2}}+2\epsilon e^{-\frac{k(k+2)}{2}(\frac{a}{\sigma})^2-\frac{2k}{k+2}}<1\\
    \Leftrightarrow\quad & (1-2\epsilon)[e^{(\frac{a}{\sigma})^2\frac{1}{2}+\frac{2}{k+2}}-1]<2\epsilon [1-e^{-\frac{k(k+2)}{2}(\frac{a}{\sigma})^2-\frac{2k}{k+2}}]\\
    \Leftrightarrow\quad & \frac{1}{2\epsilon}-1< \frac{1-e^{-\frac{k(k+2)}{2}(\frac{a}{\sigma})^2-\frac{2k}{k+2}}}{e^{(\frac{a}{\sigma})^2\frac{1}{2}+\frac{2}{k+2}}-1}\\
    \Leftrightarrow\quad & \frac{1}{2\epsilon}< \frac{e^{(\frac{a}{\sigma})^2\frac{1}{2}+\frac{2}{k+2}}-e^{-\frac{k(k+2)}{2}(\frac{a}{\sigma})^2-\frac{2k}{k+2}}}{e^{(\frac{a}{\sigma})^2\frac{1}{2}+\frac{2}{k+2}}-1}\\
    \Leftrightarrow\quad & \frac{1}{2\epsilon}< \frac{\tau l-\tau^{-k(k+2)}l^{-k}}{\tau l-1} \quad (\text{let $\tau=e^{(\frac{a}{\sigma})^2\frac{1}{2}}, l=e^{\frac{2}{k+2}}$})\\
    \Leftrightarrow\quad & \frac{1}{2\epsilon}< \tau^{-k(k+2)}l^{-k}\frac{\tau^{(k+1)^2}l^{k+1}-1}{\tau l-1}\\
    \Leftrightarrow\quad & \frac{1}{2\epsilon}< \tau^{-k(k+2)}l^{-k}\frac{\tau^{(k+1)^2}l^{k+1}-1}{\tau^{k+1}l-1}\frac{\tau^{k+1}l-1}{\tau l-1}\\
    \Leftrightarrow\quad & \frac{1}{2\epsilon}< \tau^{-k(k+1)}l^{-k}(\Sigma^k_{i=0}(\tau^{k+1}l)^i)\frac{\tau^{k+1}l-1}{\tau l-1}\tau^{-k}\\
    \Leftrightarrow\quad & \frac{1}{2\epsilon}< (\Sigma^k_{i=0}(\tau^{k+1}l)^{-i})\frac{\tau l-\tau^{-k}}{\tau l-1}\\
    \Leftarrow\quad & \frac{1}{2\epsilon}\leq \Sigma^k_{i=0}(\tau^{k+1}l)^{-i} \leq (k+1)(\tau^{k+1}l)^{-k}\\
    \Leftrightarrow\quad & 0<ln(\tau)\leq\frac{ln(2\epsilon(k+1))-k ln(l)}{k(k+1)}=\frac{ln(2\epsilon(k+1))-\frac{2k}{k+2}}{k(k+1)}\\
    \Leftarrow\quad & (\frac{a}{\sigma})^2\frac{1}{2}\leq\frac{ln(2\epsilon(k+1))-\frac{2k}{k+2}}{k(k+1)},\ k>\frac{e^2}{\epsilon}-1\\
    \Leftrightarrow\quad & \sigma\geq a\sqrt{\frac{k(k+1)}{2ln(2\epsilon(k+1))-\frac{2k}{k+2}}},\ k>\frac{e^2}{\epsilon}-1.
\end{aligned}
\end{equation*}
These conclude our proof.
\end{proof}
\end{Theorem}

\newpage
\section{Definitions for ``THEORETICAL CHARACTERIZATION OF THE SHRINKING PHENOMENON''}
\subsection{Definitions}
\setcounter{definition}{0}
\begin{definition}[Smoothed]
\label{s_def:smoothing}
If we use $f$ to denote an original neural network function with outputs in the simplex $\Delta^c= \{z \in \rrr^c \mid \sum_{i=1}^c z_i = 1, \, 0 \leq z_i \leq 1, \, \forall i\}$, then its smoothed counterpart defined on $d$-dimensional inputs $x\in\mathbb{R}^d$ is defined by
\begin{align*}
    f_{\text{smooth}}(x) &= \int_{x' \in \rrr^d} f(x') p(x')dx',
\end{align*}
where $p(x')$ is the probability density function of the filter.
\end{definition}

\begin{definition}[Gaussian smoothing]
\label{s_def:gaussiansmoothing}
If $p(x')$ is the probability density function of a normally-distributed random variable with an expected value $x$ and standard deviation $\sigma$, then we call $f_{smooth}$ a Gaussian-smoothed function and denote it by $f_\sigma$.
\end{definition}

\begin{definition}[Bounded Decision Regions]
\label{s_def:bounded}
If the decision region (disconnected or connected) of class 1 data is a bounded set in the Euclidean space (can be bounded by a ball of finite radius), then we call these decision regions bounded decision regions. 
\end{definition}

\begin{definition}[Shrinking of Bounded Decision Regions]
\label{s_def:shrinkg_bnded}
A bounded decision region is distinguished as shrinked after applying smoothing filters if the radius $R_\sigma$ of $S_{\db_\sigma}$ is rigorously smaller than the radius $R$ of $S_{\db}$, \textit{i.e. $R_\sigma<R$}, where $S_{\db}$ and $S_{\db_\sigma}$ are the smallest balls containing the original decision region and the smoothed decision region, respectively.
\end{definition}

\begin{definition}[Unbounded Decision Regions]
If for any ball there exists at least one point in the decision regions that reside outside the ball, then we call these decision regions unbounded decision regions.
\end{definition}

\begin{definition}[Semi-bounded Decision Regions]
\label{s_def:semibounded}
For an unbounded decision region, if there exists any half-space $\mathcal{H}$ (decided by a hyperplane) that contains the unbounded decision region, then we call it semi-bounded decision region. We say a semi-bounded decision region is bounded in $v$-direction if there $\exists k\in\mathbb{R}/\infty$ such that for $\forall x\in\db$, $v^{T}x<k$.
\end{definition}

\begin{definition}[Shrinking of Semi-bounded Decision Regions]
A semi-bounded decision region bounded in $v$-direction is distinguished as shrinked along the direction after applying smoothing filters if the upper bound of projections of the decision region onto direction $v$ shrinks, \textit{i.e.} $\Upsilon^v_{\db_\sigma}<\Upsilon^v_{\db}$, where $\Upsilon^v_{\db}=\max_{x\in\db} v^{T}x, \Upsilon^v_{\db_\sigma}=\max_{x\in\db_\sigma} v^{T}x$.
\end{definition}

\begin{definition}[$\theta,v$-Bounding Cone for a Decision Region]
A $\theta,v$ cone is defined as a right circular cone $\mathcal{C}$ with axis along $-v$ and aperture $2\theta$. Then we define the $\theta,v$-bounding cone $\cone$ for $\db$ as the $\theta,v$ cone that has the smallest projection on $v$ and contains $\db$, \textit{i.e.}, $\cone = \argmin_{\db \subseteq \mathcal{C}_{\theta,v}} \Upsilon^v_{\mathcal{C}_{\theta,v}}$.
\end{definition}

\newpage

\section{Proofs for ``THEORETICAL CHARACTERIZATION OF THE SHRINKING PHENOMENON''}
\subsection{Proofs for Sec 4.1}
\begin{Lemma}
\label{s_lemma:1}
    For any two original decision regions $A, B$, if we have that $A \subset B$, then we also have that $A_\sigma \subset B_\sigma$, where $A_\sigma$ and $B_\sigma$ are the decision regions of the Gaussian-smoothed functions.
\begin{proof}
Recalling that decision regions $A_\sigma$ and $B_\sigma$ satisfy $D_\sigma=\{x\in\mathbb{R}^d|f^D_\sigma(x)_1\geq \frac{1}{c}\}$ for $D=A,B$. Therefore for $\forall x\in A_\sigma$, we have $f^A_\sigma(x)\geq \frac{1}{c}$. And 
\begin{align*}
    f_\sigma^B(x)_1 &= \int_{x' \in \rrr^d} f^B(x')_1 p(x')dx' = \int_{x' \in \rrr^d} \mathbb{1}_{x' \in B} p(x')dx'\\
    &= \int_{x' \in B} p(x') dx' > \int_{x' \in A} p(x')dx'\\
    &= \int_{x' \in \rrr^d} \mathbb{1}_{x' \in A} p(x')dx' = \int_{x' \in \rrr^d} f^A(x')_1 p(x')dx'\\
    &= f^A_\sigma(x)_1\geq \frac{1}{c},
\end{align*}
implying $x\in B_\sigma$. That said, we have that if $x \in A_\sigma$, then $x \in B_\sigma$, making $A_\sigma \subseteq B_\sigma$.
\end{proof}
\end{Lemma} 

\begin{Corollary}
\label{s_cor:1}
    The smallest ball $S_{\db_\sigma}$ containing the smoothed decision region is contained within the smoothed version of $S_\db$, \textit{i.e.} $S_{\db_\sigma} \subseteq (S_\db)_\sigma$.
\begin{proof}
    As we have $\db \subseteq S_\db$, from Lemma~\ref{s_lemma:1} we get $\db_\sigma \subseteq (S_\db)_\sigma$. Then by isotropy we have that $(S_\db)_\sigma$ is also a ball centered at the same point as $S_\db$. As $S_{\db_\sigma}$ is the smallest ball containing $\db_\sigma$, we have that $S_{\db_\sigma} \subseteq (S_\db)_\sigma$.
\end{proof}
\end{Corollary}

\noindent We also need another important definition for the coming theorem, the regularized Gamma function: 
\begin{definition}[Regularized Gamma Function]
\label{s_def:regularizedgamma}
The lower regularized gamma functions $Q(s,x)$ is defined by
\begin{align*}
   Q(s,x)=\frac{\int^x_0 t^{s-1}e^{-t}dt}{\int^\infty_0 t^{s-1}e^{-t}dt}.
\end{align*}
\end{definition}
\noindent
Moreover, it is well-known that
$$Q\bigg(\frac{d}{2}, \frac{R^2}{2\sigma^2}\bigg) = \int_{x'\in \rrr^d, \|x'\|_2\leq R } (2\pi\sigma^2)^{-\frac{d}{2}}e^{\frac{x'^Tx'}{2\sigma^2}} dx'.$$
We also give a short proof of this in the proof of Theorem \ref{s_thm:2}.
\noindent For the number of dimensions $d$, we summarize the lemma based on regularized Gamma functions below. 

\begin{Lemma}
\label{s_lemma:2}
For $\forall d,c \in\mathbb{N}^+$, $Q(\frac{d}{2},\frac{d}{2c})<\frac{1}{c} $ holds.
\begin{proof}
To prove $Q(\frac{d}{2},\frac{d}{2c}) < \frac{1}{c})$, by definition~\ref{s_def:regularizedgamma}, we aim at proving $\int^\infty_0 t^{\frac{d}{2}-1}e^{-t}dt>c\cdot\int^{\frac{d}{2c}}_0 t^{\frac{d}{2}-1}e^{-t}dt$ ($\forall d\in\mathbb{N}^+$). For c = 1, this is clearly true as $t^{x-1}e^{-t}\geq 0$ is true for $t\geq 0$. Then we show it also holds for $c\geq 2$.

Let $g(t)=t^{x-1}e^{-t}$, we have $g'(t)=t^{x-2}e^{-t}(x-1-t)$. Therefore $g(t)$ is increasing when $t\leq x-1$ and decreasing when $t>x-1$. Thus, giving us two equations 
\begin{align*}
    \int^x_{\frac{x}{c}} t^{x-1}e^{-t}dt &>\min\{x^{x-1}e^{-x}, {(\frac{x}{c})}^{x-1}e^{-\frac{x}{c}}\}\frac{(c-1)x}{c} \\
    \frac{x}{c}{(\frac{x}{c})}^{x-1}e^{-\frac{x}{c}} &> \int^{\frac{x}{c}}_0 t^{x-1}e^{-t}dt
\end{align*}
So, we see that for any $x,c$ if we have $x^{x-1}e^{-x} \geq {(\frac{x}{c})}^{x-1}e^{-\frac{x}{c}}$ then
$\int^x_{\frac{x}{c}} t^{x-1}e^{-t}dt > (c-1)\cdot \int^{\frac{x}{c}}_0 t^{x-1}e^{-t}dt \Leftrightarrow 
\int^x_0 t^{x-1}e^{-t}dt > c \cdot \int^{\frac{x}{c}}_0 t^{x-1}e^{-t}dt.$
Using $t^{x-1}e^{-t}\geq 0, \forall{x} \int^\infty_0 t^{x-1}e^{-t}dt \geq \int^x_0 t^{x-1}e^{-t}dt$. So, we have $\int^\infty_0 t^{x-1}e^{-t}dt > c \cdot \int^{\frac{x}{c}}_0 t^{x-1}e^{-t}dt$ as needed. So, for any $x,c$ it is sufficient to show
$$x^{x-1}e^{-x} \geq {(\frac{x}{c})}^{x-1}e^{-\frac{x}{c}}$$
in order to prove $\int^\infty_0 t^{x-1}e^{-t}dt > c \cdot \int^{\frac{x}{c}}_0 t^{x-1}e^{-t}dt$. The inequality can be re-written as $(x-1) \log(c) > \frac{c-1}{c}x$ or $(1 - \frac{1}{x}) > (1 - \frac{1}{c})\frac{1}{\log(c)}$. We observe that $(1 - \frac{1}{c})\frac{1}{\log(c)}$ is a decreasing function of $c$ for $c \geq 1$ and $(1 - \frac{1}{x})$ is an increasing function of $x$.\\
For $x \geq 4, c \geq 2$, we see $(1 - \frac{1}{x}) \geq 1 - \frac{1}{4} = 0.75 >  (1 - \frac{1}{2})\frac{1}{\log(2)} \geq (1 - \frac{1}{c})\frac{1}{\log(c)}$.\\
For $x \geq \frac{3}{2}, c \geq 20$, we have $(1 - \frac{1}{x}) \geq 1 - \frac{2}{3} >  (1 - \frac{1}{20})\frac{1}{\log(20)} \geq (1 - \frac{1}{c})\frac{1}{\log(c)}$.\\
For $\frac{3}{2} \leq x < 4 \rightarrow 3 \leq d < 8$ and $2 \leq c < 20$, we numerically verify the values of $Q(\frac{d}{2}, \frac{d}{2c})$ to see the inequality is satisfied.\\
Thus, for $d \geq 3, c \geq 2$ we have the inequality. \\
\\
For $d=2$, we have $Q(\frac{d}{2}, \frac{d}{2c}) = Q(1, \frac{1}{c})$. This has a closed form solution $Q(1, x) = 1 - e^{-x}$. So, we need to show that for $c\geq 2$ $1 - e^{-\frac{1}{c}} < \frac{1}{c}$ or $e^{\frac{1}{c}} < \frac{c}{c - 1}$ or $\frac{1}{c} < \log(1 + \frac{1}{c-1})$. But we know that for $x > -1, x \neq 0$, $\log(1 + x) > \frac{x}{x+1}$, so $\log( 1 + \frac{1}{c-1}) > \frac{\frac{1}{c-1}}{1 + \frac{1}{c-1}} = \frac{1}{c}$ which concludes the proof for $d = 2, c \geq 2$.
\end{proof}
\end{Lemma}

\begin{Theorem}
\label{s_thm:1}
A bounded decision region shrinks after applying Gaussian smoothing filters with large $\sigma$, \textit{i.e.} if $\sigma > \frac{R\sqrt{c}}{\sqrt{2(d-1)}} $, then $R_\sigma<R$, where $R$ and $R_\sigma$ are the radii of $S_{\db}$ and $S_{\db_\sigma}$, the smallest balls bounding the original decision region and the smoothed decision region, respectively.
\begin{proof}
Considering the ball $S_\db$, we see that from Corollary \ref{s_cor:1}, $\db_\sigma \subseteq (S_\db)_\sigma$. Thus, we see that by the definition of radius $R_{\db_\sigma} \leq R_{(S_\db)_\sigma}$. It is sufficient to show that for large $\sigma$, $R_{(S_\db)_\sigma} < R_{S_\db}$. 
\noindent
Then we observe that due to the isotropic nature of Gaussian smoothing, $(S_\db)_\sigma$ is also a sphere concentric to $S_\db$. So, it is sufficient to show that for a point $x$ at distance $R_{S_\db}$ from the center $x_0$ of the sphere, $f_\sigma(x)_1  < \frac{1}{c}$.\\
Without loss of generality consider $\db$ to be the origin-centered sphere of radius $R$ and $x = [0, \ldots, 0, R]^T$. It is sufficient to show for large $\sigma$ $f_\sigma(x)_1 < \frac{1}{c}$. By definition~\ref{s_def:gaussiansmoothing}, we have
\begin{align}
    f_\sigma(x)_1 &=\nonumber \int_{x' \in \rrr^d} f(x')_1 p(x')dx'\\ 
    &=\nonumber  \int_{\|x'\|_2\leq R } (2\pi)^{-\frac{d}{2}}|\Sigma|^{-\frac{1}{2}}e^{-\frac{1}{2}(x'-x)^T\Sigma^{-1} (x'-x)} dx'\\ 
    &=  \int_{\|x'\|_2\leq R } (2\pi\sigma^2)^{-\frac{d}{2}}e^{-\frac{(x'-x)^T (x'-x)}{2\sigma^2}} dx'.\label{supeqn:smoothing}
\end{align}

\noindent
Then substituting the value of $x$, we get the equation.
\begin{align*}
    f_\sigma(x)_1 &=  \int_{\|x'\|_2\leq R } (2\pi\sigma^2)^{-\frac{d}{2}}e^{\frac{\sum_{i=1}^{d-1} x_i'^2 + (x_d' - R)^2}{2\sigma^2}} dx' \\
    &= \int_{-R}^{R}\int_{\sum^{d-1}_{k=1} {x_k'}^2\leq R^2-{x_d'}^2}(2\pi\sigma^2)^{-\frac{d}{2}} e^{-\frac{\sum^{d-1}_{k=1} (x_k'-x_k)^2}{2\sigma^2}} dx_1'\ldots dx_{d-1}' e^{-\frac{ (x_d'-x_d)^2}{2\sigma^2}}dx_d' \\
    &< \int_{-R}^{R}\int_{\sum^{d-1}_{k=1} {x_k'}^2\leq R^2} (2\pi\sigma^2)^{-\frac{d}{2}} e^{-\frac{\sum^{d-1}_{k=1} (x_k'-x_k)^2}{2\sigma^2}} dx_1'\ldots dx_{d-1}' e^{-\frac{ (x_d'-x_d)^2}{2\sigma^2}}dx_d' \\
    &= (\int_{-R}^{R}(2\pi\sigma^2)^{-\frac{1}{2}}e^{-\frac{ (x_d'-x_d)^2}{2\sigma^2}}dx_d')(\int_{\sum^{d-1}_{k=1} {x_k'}^2\leq R^2} (2\pi\sigma^2)^{-\frac{d-1}{2}}e^{-\frac{\sum^{d-1}_{k=1} (x_k'-x_k)^2}{2\sigma^2}} dx_1'\ldots dx_{d-1}') \\
    &= (\Phi(\frac{2R}{\sigma}) - \Phi(0)) \cdot Q(\frac{d-1}{2}, \frac{R^2}{2\sigma^2})\\
    &< \frac{1}{2}\cdot Q(\frac{d-1}{2}, \frac{R^2}{2\sigma^2}).
\end{align*}
\noindent
Using Lemma \ref{s_lemma:2} we get that for $d \geq 3$, if $\frac{R^2}{2\sigma^2} \leq \frac{d-1}{c}$, then we have $\frac{1}{2}\cdot Q(\frac{d-1}{2}, \frac{R^2}{2\sigma^2}) < \frac{1}{c}$. Now, 
$\frac{R^2}{2\sigma^2} < \frac{d-1}{c}$ gives
$$\sigma > {\frac{R\sqrt{c}}{\sqrt{2(d-1)}}}. $$
\end{proof}
\end{Theorem}

For class 1 data $x$, the point at the origin has the highest probability to be classified as class 1, \textit{i.e.} $f_\sigma(x)_1 = \int_{x' \in \rrr^d} f(x') p(x')dx'  =  \int_{\|x'\|_2\leq R } (2\pi)^{-\frac{d}{2}}|\Sigma|^{-\frac{1}{2}}e^{-\frac{1}{2}(x'-x)^T\Sigma^{-1} (x'-x)} dx' \leq f_\sigma(0)_1$.

\begin{Lemma}
\label{s_thm:supp1}
Assume the decision region of class 1 data is $\{x\in\mathbb{R}^d\mid \|x\|_2\leq R\}$, the point at the origin has the highest probability to be classified as class 1 by the gaussian-smoothed classifier $f_\sigma$, \textit{i.e.} $f_\sigma(x)_1 \leq f_\sigma(0)_1$.
\begin{proof}
We do the proof by mathematical induction and begin by giving $d=1$ case. For $\forall R>0$ and $d=1$, Equation~\eqref{supeqn:smoothing} reduces to
\begin{align*}
    f_\sigma(x)_1 &= \int^R_{-R} (2\pi\sigma^2)^{-\frac{1}{2}}e^{-\frac{(x'-x)^2}{2\sigma^2}} dx'\\
    &\xlongequal{a=x'-x} \int^{R-x}_{-R-x} (2\pi\sigma^2)^{-\frac{1}{2}}e^{-\frac{a^2}{2\sigma^2}} da\\
    f'_\sigma(x)_1 &= -(2\pi\sigma^2)^{-\frac{1}{2}}e^{-\frac{(R-x)^2}{2\sigma^2}}-(-1)(2\pi\sigma^2)^{-\frac{1}{2}}e^{-\frac{(-R-x)^2}{2\sigma^2}}
\end{align*}
and $f'_\sigma(x)_1$ equals to zero only when $x=0$. Now suppose the conclusion holds for $d-1$ dimensional case, then when $x\in\mathbb{R}^d$ we scale $f_\sigma(x)_1$ by $(2\pi\sigma^2)^{\frac{d}{2}}$ and obtain
\begin{align*}
    &\int_{\|x'\|_2\leq R } e^{-\frac{(x'-x)^T (x'-x)}{2\sigma^2}} dx'\\
    =& \int_{\sum^d_{k=1} {x_k'}^2\leq R^2 } e^{-\frac{\sum^d_{k=1} (x_k'-x_k)^2}{2\sigma^2}} dx'\\
    =& \int_{-R}^{R}\int_{\sum^{d-1}_{k=1} {x_k'}^2\leq R^2-{x_d'}^2} e^{-\frac{\sum^{d-1}_{k=1} (x_k'-x_k)^2}{2\sigma^2}} dx_1'\ldots dx_{d-1}' e^{-\frac{ (x_d'-x_d)^2}{2\sigma^2}}dx_d'\\
    \leq& \int_{-R}^{R}\int_{\sum^{d-1}_{k=1} {x_k'}^2\leq R^2-{x_d'}^2} e^{-\frac{\sum^{d-1}_{k=1} {x_k'}^2}{2\sigma^2}} dx_1'\ldots dx_{d-1}' e^{-\frac{ (x_d'-x_d)^2}{2\sigma^2}}dx_d'\\
    =& \int_{\sum^d_{k=1} {x_k'}^2\leq R^2 } e^{-\frac{\sum^{d-1}_{k=1} {x_k'}^2}{2\sigma^2}} e^{-\frac{ (x_d'-x_d)^2}{2\sigma^2}}dx'\\
    =& \int_{\sum^{d-1}_{k=1} {x_k'}^2\leq R^2}\int_{{x_d'}^2\leq R^2-\sum^{d-1}_{k=1} {x_k'}^2} e^{-\frac{ (x_d'-x_d)^2}{2\sigma^2}}dx_d' e^{-\frac{\sum^{d-1}_{k=1} {x_k'}^2}{2\sigma^2}} dx_1'\ldots dx_{d-1}'\\
    \leq& \int_{\sum^{d-1}_{k=1} {x_k'}^2\leq R^2}\int_{{x_d'}^2\leq R^2-\sum^{d-1}_{k=1} {x_k'}^2} e^{-\frac{ {x_d'}^2}{2\sigma^2}}dx_d' e^{-\frac{\sum^{d-1}_{k=1} {x_k'}^2}{2\sigma^2}} dx_1'\ldots dx_{d-1}'\\
    =& \int_{\sum^d_{k=1} {x_k'}^2\leq R^2 } e^{-\frac{\sum^{d}_{k=1} {x_k'}^2}{2\sigma^2}} dx',
\end{align*}
where the first inequality comes from the assumption that the conclusion holds for $d-1$ dimensional case with equality if and only if $x_1=\ldots x_{d-1}=0$, and the second inequality comes from an one dimensional observation with equality precisely when $x_d=0$. This concludes our proof.
\end{proof}
\end{Lemma}

Since the value of $f_\sigma(0)_1$ depends on the radius $R$ of the decision region, the dimension $d$, and the smoothing factor $\sigma$, we denote $f_\sigma(0)_1$ by $q(R,d,\sigma)$, \textit{i.e.} $q(R,d,\sigma):=f_\sigma(0)_1$.

\begin{Theorem}[Vanishing Rate in the Ball-like Decision Region Case]
\label{s_thm:2}
The decision region of class 1 data vanishes at smoothing factor $\sigma_{\text{van}}>\frac{R\sqrt{c}}{\sqrt{d}}$. 

\begin{proof}
    Noticing that the surface area of a $d$-dimensional ball of radius $r$ is proportional to $r^{d-1}$, we can therefore write out the probability of the point at the origin be classified as class 1 as 
    \begin{align*}
        q(R,d,\sigma)&=\frac{\int^R_0 r^{d-1}(\frac{1}{2\pi \sigma^2})^{\frac{d}{2}}e^{-\frac{r^2}{2\sigma^2}}dr}{\int^\infty_0 r^{d-1}(\frac{1}{2\pi \sigma^2})^{\frac{d}{2}}e^{-\frac{r^2}{2\sigma^2}}dr}\\
        &=\frac{\int^R_0 r^{d-1}e^{-\frac{r^2}{2\sigma^2}}dr}{\int^\infty_0 r^{d-1}e^{-\frac{r^2}{2\sigma^2}}dr}\\
        &\xlongequal{t=\frac{r^2}{2\sigma^2}}\frac{\int^{\frac{R^2}{2\sigma^2}}_0(2\sigma^2t)^{\frac{d-1}{2}}e^{-t}\sigma^2(2\sigma^2t)^{-\frac{1}{2}}dt}{\int^\infty_0(2\sigma^2t)^{\frac{d-1}{2}}e^{-t}\sigma^2(2\sigma^2t)^{-\frac{1}{2}}dt}\\
        &=\frac{\int^{\frac{R^2}{2\sigma^2}}_0t^{\frac{d}{2}-1}e^{-t}dt}{\int^\infty_0t^{\frac{d}{2}-1}e^{-t}dt}\\
        &=Q(\frac{d}{2},\frac{R^2}{2\sigma^2}).
    \end{align*}
    Now let $\sigma=\sqrt{\frac{c}{d}}R$ yields $q(R,d,\sqrt{\frac{c}{d}}R)=Q(\frac{d}{2},\frac{d}{2c})$. By Lemma~\ref{s_lemma:2}, we then have $Q(\frac{d}{2},\frac{d}{2c}) < \frac{1}{c}$, implying the decision region of class 1 data has already vanished and making $\sigma=\sqrt{\frac{c}{d}}R$ an upper bound of the vanishing smoothing factor.
\end{proof}
\end{Theorem}

\subsection{Proofs for Sec 4.2}

\begin{Corollary}
\label{s_cor:semi2}
    As $\db \subseteq \cone$, using Lemma \ref{s_lemma:1}, we have that the smoothed decision region is contained within the smoothed version of $\cone$, \textit{i.e.} $\db_\sigma \subseteq (\cone)_\sigma$.\qed
\end{Corollary}

\begin{Lemma}
\label{s_lemma:semi_max_prob}
If the decision region of class 1 data is $\db=\{x\in\mathbb{R}^d\mid v^Tx+\|v\|\|x\|cos(\theta)\leq 0\}$, where $v=[0,\ldots,0,1]^T\in\mathbb{R}^d$ and $2\theta\in(-\pi,\pi)$, then after smoothing among the set of points $S_a$ with the same projection on $v$ the point on the axis has the highest probability of being in class 1. For $S_a = \{x \mid v^Tx = a\},$ we have $\text{argsup}_{x \in S_a} f_\sigma(x)_1 = a\cdot v$. Moreover if $a_1 > a_2$, then $f_\sigma(a_1 \cdot v)_1 < f_\sigma(a_2\cdot v)_1$.
\begin{proof}
For the first part of the proof consider the set of points $S_a = \{x \mid v^Tx = a\}$. For any point $x$ is $S_a$, we see that
\begin{align*}
    f_\sigma(x)_1 &= \int_{x' \in \rrr^d} f(x')_1 p(x')dx'\\ 
    &= \int_{x'_d+\|x'\|cos(\theta)\leq 0} (2\pi\sigma^2)^{-\frac{d}{2}}e^{-\frac{(x'-x)^T (x'-x)}{2\sigma^2}} dx'\\
    &= (2\pi\sigma^2)^{-\frac{d}{2}}\int_{-\infty}^0\int_{\Sigma^{d-1}_{k=1}{x_k'}^2\leq tan^2(\theta){x'_d}^2} e^{-\frac{\sum^{d-1}_{k=1} (x_k'-x_k)^2}{2\sigma^2}} dx_1'\ldots dx_{d-1}' e^{-\frac{ (x_d'-a)^2}{2\sigma^2}}dx_d'\\
    &\leq (2\pi\sigma^2)^{-\frac{d}{2}}\int_{-\infty}^0\int_{\Sigma^{d-1}_{k=1}{x_k'}^2\leq tan^2(\theta){x'_d}^2} e^{-\frac{\sum^{d-1}_{k=1} {x_k'}^2}{2\sigma^2}} dx_1'\ldots dx_{d-1}' e^{-\frac{ (x_d'-a)^2}{2\sigma^2}}dx_d' \\
    &= f_\sigma(av)_1.
\end{align*}
where the inequality comes from Theorem~\ref{s_thm:supp1} with equality iff $x_1=\ldots x_{d-1}=0$, \textit{i.e.} $x=[0,\ldots,0,a]\in\mathcal{V}$. Now for the second part of the proof, let $x_1 = a_1v$, $x_2 = a_2v$ such that $a_1 > a_2$. Then
\begin{align*}
    f_\sigma(x_1)_1 &= (2\pi\sigma^2)^{-\frac{d}{2}}\int_{-\infty}^{-a_1}\int_{\Sigma^{d-1}_{k=1}{x_k'}^2\leq tan^2(\theta){(x'_d + a_1)}^2} e^{-\frac{\sum^{d-1}_{k=1} {x_k'}^2}{2\sigma^2}} dx_1'\ldots dx_{d-1}' e^{-\frac{ x_d'^2}{2\sigma^2}}dx_d' \\
    & \quad \text{As } a_1 + x'_d \leq 0, (a_1 + x'_d)^2 < (a_2 + x'_d)^2 \\
    &< (2\pi\sigma^2)^{-\frac{d}{2}}\int_{-\infty}^{-a_1}\int_{\Sigma^{d-1}_{k=1}{x_k'}^2\leq tan^2(\theta){(x'_d + a_2)}^2} e^{-\frac{\sum^{d-1}_{k=1} {x_k'}^2}{2\sigma^2}} dx_1'\ldots dx_{d-1}' e^{-\frac{ x_d'^2}{2\sigma^2}}dx_d' \\
    &< (2\pi\sigma^2)^{-\frac{d}{2}}\int_{-\infty}^{-a_2}\int_{\Sigma^{d-1}_{k=1}{x_k'}^2\leq tan^2(\theta){(x'_d + a_2)}^2} e^{-\frac{\sum^{d-1}_{k=1} {x_k'}^2}{2\sigma^2}} dx_1'\ldots dx_{d-1}' e^{-\frac{ x_d'^2}{2\sigma^2}}dx_d' \\
    &= f_\sigma(x_2)_1.
\end{align*}

\end{proof}
\end{Lemma}

\begin{Lemma}
\label{s_lemma:cdfineq2}
$\forall a>0, k\geq 1$, $\frac{\Phi(-a)}{\Phi(-ka)} \geq e^{\frac{(k^2-1)a^2}{2}}$.
\begin{proof}
Consider the function $h(x) = \frac{\sqrt{2\pi}\Phi(-x)}{e^{-x^2/2}}$ and we will show in the following that it is strictly decreasing for $x > 0$. Alternatively, we take the derivative \textit{w.r.t.} $x$,
$$ \frac{d}{dx}h(x) = \frac{\sqrt{2\pi}x\Phi(-x)}{e^{-x^2/2}} - 1,$$
and show that it is negative for $x > 0$. Since $e^{-x^2/2}>0$, it is sufficient to show that $\sqrt{2\pi}x\Phi(-x)-e^{-x^2/2}<0$.
Combining that 1) $\sqrt{2\pi}x\Phi(-x)-e^{-x^2/2}$ is increasing as
\begin{align*}
    \frac{d}{dx} \bigg(x\Phi(-x) - \frac{e^{-x^2/2}}{\sqrt{2\pi}}\bigg)  &= \Phi(-x) - \frac{xe^{-x^2/2}}{\sqrt{2\pi}} - \frac{-xe^{-x^2/2}}{\sqrt{2\pi}}\\
    &= \Phi(-x) > 0
\end{align*}
and 2) $\sqrt{2\pi}x\Phi(-x)-e^{-x^2/2}\rightarrow 0$ when $x \rightarrow \infty$, we have that $\sqrt{2\pi}x\Phi(-x)-e^{-x^2/2}<0$.
As $h(x)$ is strictly decreasing we have that for any $a > 0$ and $k > 1$, $ka > a$. Thus,
$$ \frac{\sqrt{2\pi}\Phi(-a)}{e^{-a^2/2}} > \frac{\sqrt{2\pi}\Phi(-ka)}{e^{-(ka)^2/2}}.$$
Rearranging the terms gives the inequality.
\end{proof}
\end{Lemma}

\begin{Theorem}
\label{s_thm:semi1}
A semi-bounded decision region that has a narrow bounding cone shrinks along $v$-direction after applying Gaussian smoothing filters with high $\sigma$, \textit{i.e.} if the region admits a bounding cone $\cone$ with $\tan(\theta)< \sqrt{\frac{(d-1)}{2c\log(c-1)}}$, then for $\sigma > (\Upsilon^v_\cone - \Upsilon^v_\db)\tan(\theta)\sqrt{\frac{c}{d-1}}\cdot\frac{2(d-1)}{(d-1) - 2\tan^2(\theta)c\log(c-1)}$, $\Upsilon^v_{\db_\sigma}<\Upsilon^v_{\db}$.
\begin{proof}
    In this derivation we assume without loss of generality, $v = [0, \ldots, 0,1]^T \in \rrr^d$ (It is always possible to orient the axis to make this happen). From Corollary \ref{s_cor:semi2}, we can see that $\db_\sigma \subseteq (\cone)_\sigma$ which gives us $\Upsilon^v_{\db_\sigma} = \max_{x\in\db_\sigma} v^{T}x \leq \max_{x\in(\cone)_\sigma} v^{T}x = \Upsilon^v_{(\cone)_\sigma}$. Then to show that $\Upsilon^v_{\db_\sigma}<\Upsilon^v_{\db}$ it is sufficient to show that $\Upsilon^v_{(\cone)_\sigma} < \Upsilon^v_{\db}$. \\
    We observe that we only need to check the point $x$ on the axis of the cone at distance  $\Upsilon^v_\cone - \Upsilon^v_\db$ from the tip $x_0$ of the cone, \textit{i.e.}, $x = x_0 - (\Upsilon^v_\cone - \Upsilon^v_\db)v$. If $x$ is not classified as Class 1 then by Lemma \ref{s_lemma:semi_max_prob}, we have that
    \begin{align*}
        \Upsilon^v_{(\cone)_\sigma} &< v^Tx = v^T(x_0 - (\Upsilon^v_\cone - \Upsilon^v_\db)v) \\
                                    &= v^Tx_0 - (\Upsilon^v_\cone - \Upsilon^v_\db) v^Tv \\           
                                    &= \Upsilon^v_\cone - (\Upsilon^v_\cone - \Upsilon^v_\db) = \Upsilon^v_\db 
    \end{align*}
    From the above argument and the definition of the decision boundary we see that if $f_\sigma(x)_1 < \frac{1}{c}$, then $\Upsilon^v_{\db_\sigma}<\Upsilon^v_{\db}$. Without loss of generality we let $x_0$ be the origin. By definition~\ref{s_def:gaussiansmoothing}, we have
    \begin{align*}
        f_\sigma(x)_1 &= \int_{x' \in \rrr^d} f(x')_1 p(x')dx'\\ 
    &= \int_{x_d'+\|x'\|cos(\theta)\leq 0} (2\pi\sigma^2)^{-\frac{d}{2}}e^{-\frac{(x'-x)^T (x'-x)}{2\sigma^2}} dx'\\
    &= (2\pi\sigma^2)^{-\frac{d}{2}}\int_{-\infty}^0\int_{\Sigma^{d-1}_{k=1}{x_k'}^2\leq tan^2(\theta){x_d'}^2} e^{-\frac{\sum^{d-1}_{k=1} (x_k'-x_k)^2}{2\sigma^2}} dx_1'\ldots dx_{d-1}' e^{-\frac{ (x_d'-x_d)^2}{2\sigma^2}}dx_d'\\
    &= (2\pi\sigma^2)^{-\frac{d}{2}}\int_{-\infty}^0\int_{\Sigma^{d-1}_{k=1}{x_k'}^2\leq tan^2(\theta){x_d'}^2} e^{-\frac{\sum^{d-1}_{k=1} {x_k'}^2}{2\sigma^2}} dx_1'\ldots dx_{d-1}' e^{-\frac{ (x_d'-x_d)^2}{2\sigma^2}}dx_d' \\
    &= (2\pi\sigma^2)^{-\frac{1}{2}} \int_{-\infty}^0 q(|x'_d\tan(\theta)|, d-1, \sigma)e^{-\frac{ (x_d'-x_d)^2}{2\sigma^2}} dx_d' \\
    &= (2\pi\sigma^2)^{-\frac{1}{2}} \int_{-\infty}^0 Q(\frac{d-1}{2},\frac{tan^2(\theta){x_d'}^2}{2\sigma^2})e^{-\frac{ (x_d'-x_d)^2}{2\sigma^2}} dx_d' \\ 
    &\text{Substitute } X_d = \frac{x_d}{\sigma}, X'_d = \frac{x'_d}{\sigma}\\
    &= (2\pi)^{-\frac{1}{2}} \int_{-\infty}^0 Q(\frac{d-1}{2},\frac{tan^2(\theta){X_d'}^2}{2})e^{-\frac{ (X_d'-X_d)^2}{2}} dX_d' \\
    &\text{Let } M \leq \sqrt{\frac{d-1}{c\tan^2(\theta)}}, k = \frac{M}{X_d}\\
    &= (2\pi)^{-\frac{1}{2}} \int_{M}^0 Q(\frac{d-1}{2},\frac{tan^2(\theta){X_d'}^2}{2})e^{-\frac{ (X_d'-X_d)^2}{2}} dX_d' \\ 
    &+ (2\pi)^{-\frac{1}{2}}\int_{-\infty}^{M} Q(\frac{d-1}{2},\frac{tan^2(\theta){X_d'}^2}{2})e^{-\frac{ (X_d'-X_d)^2}{2}} dX_d' \\
    &< (\Phi(-X_d) - \Phi(M - X_d))Q(\frac{d-1}{2}, \frac{\tan^2(\theta) M^2}{2}) +\Phi(M-X_d) \\
    &< \frac{\Phi(-X_d) - \Phi(M-X_d)}{c} + \Phi(M-X_d)\\
    &= \frac{1}{c} + \frac{(c-1) \Phi((k-1)X_d) - \Phi(X_d)}{c}.
    \end{align*}
Then we see that using Lemma \ref{s_lemma:cdfineq2}, we see that we see that if $e^{\frac{X_d^2((k-1)^2 - 1)}{2}} \geq c-1$ then
$(c-1)\Phi((k-1)X_d) \leq \Phi(X_d)$. So, we need to satisfy the inequalities for some $k$:
$$\sqrt{\frac{2\log(c-1)}{(k-1)^2 - 1}}\leq -X_d \leq \sqrt{\frac{d-1}{k^2c\tan^2(\theta)}}.$$
This is only possible if for some $k$, we have $\sqrt{\frac{2\log(c-1)}{(k-1)^2 - 1}}\leq \sqrt{\frac{d-1}{k^2c\tan^2(\theta)}}$ or $\tan(\theta) \leq \sqrt{\frac{d-1}{2c\log(c-1)}}.\sqrt{1 - \frac{2}{k}}$. So, we need that 
$$\tan(\theta) < \sqrt{\frac{d-1}{2c\log(c-1)}}.$$
Then we see that giving the cone is narrow enough, we have the required shrinking if we have $X_d$ satisfies the inequalities for some k. So, we see that if we have $-X_d = \sqrt{\frac{d-1}{k^2c\tan^2(\theta)}}$ for some $k$ such that $\tan(\theta) \leq \sqrt{\frac{d-1}{2c\log(c-1)}}.\sqrt{1 - \frac{2}{k}}$ is satisfied. So, we need that
$\frac{-x_d}{\sigma} = \sqrt{\frac{d-1}{k^2c\tan^2(\theta)}}$ for some suitable $k$. Thus we need $\sigma = -x_d\tan(\theta)\sqrt{\frac{c}{d-1}} k$ for some suitable $k$. Including the constraint on $k$ and substituting the value for $x_d$, we get that shrinking always happens for 
$$\sigma  \geq (\Upsilon^v_\cone - \Upsilon^v_\db)\tan(\theta)\sqrt{\frac{c}{d-1}}\cdot\frac{2(d-1)}{(d-1) - 2\tan^2(\theta)c\log(c-1)}.$$
\end{proof}
\end{Theorem}

\begin{Theorem}
\label{s_thm:semi2}
The shrinkage of class 1 decision region is proportional to the smoothing factor, \textit{i.e.} \mbox{$\Upsilon^v_{\mathcal{D}}-\Upsilon^v_{\mathcal{D}_{\sigma}}\propto\sigma$}. 
\begin{proof}
   In this case we assume a cone-like decision region which can be represented as $\db = \{x\in\mathbb{R}^d\mid v^Tx+\|v\|\|x\|cos(\theta)\leq 0\}$ with $v= [0, \ldots, 0,1]^T$ without loss of generality. By Lemma \ref{s_lemma:semi_max_prob}, we see that in order to get bounds on $\Upsilon^v_{\db_\sigma}$, we only need to analyze the value of $f_\sigma(x)_1$ for points $x$ along the axis of the cone. Then we see that for a general point $x = av$ along the axis of the cone, using the same ideas as in proof of Theorem \ref{s_thm:semi1}, we have
   
   \begin{align*}
    f_\sigma(x)_1 &= \int_{x' \in \rrr^d} f(x')_1 p(x')dx'\\ 
    &= (2\pi\sigma^2)^{-\frac{d}{2}}\int_{-\infty}^0\int_{\Sigma^{d-1}_{k=1}{x_k'}^2\leq tan^2(\theta){x_d'}^2} e^{-\frac{\sum^{d-1}_{k=1} {x_k'}^2}{2\sigma^2}} dx_1'\ldots dx_{d-1}' e^{-\frac{ (x_d'-a)^2}{2\sigma^2}}dx_d' \\
    &= (2\pi\sigma^2)^{-\frac{1}{2}} \int_{-\infty}^0 Q(\frac{d-1}{2},\frac{tan^2(\theta){x_d'}^2}{2\sigma^2})e^{-\frac{ (x_d'- a)^2}{2\sigma^2}} dx_d' \\ 
    &\text{Substitute } A = \frac{a}{\sigma}, x'_d = \frac{x'_d}{\sigma}\\
    &= (2\pi)^{-\frac{1}{2}} \int_{-\infty}^0 Q(\frac{d-1}{2},\frac{tan^2(\theta){x_d'}^2}{2})e^{-\frac{ (x_d'-A)^2}{2}} dx_d' \\
    &= f_1(Av)_1 = f_1(\frac{1}{\sigma}x)_1 .
   \end{align*}
   \noindent Using the equation above we see that for smoothing by a general $\sigma$, 
   \begin{align*}
       \Upsilon_{\db_\sigma} &= \sup_{x \mid f_\sigma(x) \geq \frac{1}{c}} v^Tx = \sup_{x \mid f_1(\frac{1}{\sigma}x) \geq \frac{1}{c}} v^Tx = \sup_{x' \mid f_1(x') \geq \frac{1}{c}} v^T(\sigma x') \\ 
       &= \sigma \sup_{x' \mid f_1(x') \geq \frac{1}{c}} v^Tx'  = \sigma \Upsilon_{\db_1}.
   \end{align*}
   In this case we have $\Upsilon_\db = 0$ by construction, so $\Upsilon_\db - \Upsilon_{\db_\sigma} = 0 - \sigma\Upsilon_{\db_1} = \sigma \cdot (- \Upsilon_{\db_1}) \propto \sigma$.
\end{proof}
\end{Theorem}
With the above Theorem~\ref{s_thm:semi2}, we can fix the smoothing factor to $\sigma=1$ and further obtain a lower bound of the shrinking rate \textit{w.r.t} $c$, $\theta$, and $d$:
\begin{Theorem}
\label{s_thm:semi3}
The shrinking rate of class 1 decision region is at least $\sqrt{\frac{d-1}{c\tan^2(\theta)}}\cdot\frac{(d-1) - 2\tan^2(\theta)c\log(c-1)}{2(d-1)}$, \textit{i.e.} $\frac{\Upsilon^v_{\db_\sigma}-\Upsilon^v_{\db_{\sigma+\delta}}}{\delta}>\sqrt{\frac{d-1}{c\tan^2(\theta)}}\cdot\frac{(d-1) - 2\tan^2(\theta)c\log(c-1)}{2(d-1)}$.
\begin{proof}
    As in Theorem \ref{s_thm:semi2}, we assume a cone at origin along $v = [0,\ldots, 0, 1]^T$ given by $\db = \{x\in\mathbb{R}^d\mid v^Tx+\|v\|\|x\|cos(\theta)\leq 0\}$.
    Following the same proof idea as Theorem \ref{s_thm:semi2}, we see that the rate is given by the value $- \Upsilon_{\db_1}$. So, we try to get a bound on the value of $ - \Upsilon_{\db_1}$. To establish a lower bound we show that for the point $x = av$, $f_1(x)_1 < \frac{1}{c}$. Then by Lemma \ref{s_lemma:semi_max_prob} we have $\Upsilon_{\db_1} < a$ or $-\Upsilon_{\db_1} > -a$. \\
    Using the same procedure as in the proof of Theorem \ref{s_thm:semi1}, we get that if $x$ satisfies the two inequalities 
    $$\sqrt{\frac{2\log(c-1)}{(k-1)^2 - 1}}\leq -v^Tx \leq \sqrt{\frac{d-1}{k^2c\tan^2(\theta)}} $$
    for suitable real $k$, then we have $f_1(x)_1 < \frac{1}{c}$.
    So, we need $v^Tx = -\sqrt{\frac{d-1}{k^2c\tan^2(\theta)}}$ for some $k$ such that $\sqrt{\frac{2\log(c-1)}{(k-1)^2 - 1}}\leq x \leq \sqrt{\frac{d-1}{k^2c\tan^2(\theta)}}$. The constraint on $k$ can be re-written as $k \geq \frac{2(d-1)}{(d-1) - 2\tan^2(\theta)c\log(c-1)}$. Taking $k$ to be lower bound, we get that for
    $$- a = - v^Tx = \sqrt{\frac{d-1}{c\tan^2(\theta)}}\cdot\frac{(d-1) - 2\tan^2(\theta)c\log(c-1)}{2(d-1)}$$
    \noindent $f_1(x)_1 \leq \frac{1}{c}$. So, we get that the rate is $-\Upsilon_{\db_1} \geq -a \geq \sqrt{\frac{d-1}{c\tan^2(\theta)}}\cdot\frac{(d-1) - 2\tan^2(\theta)c\log(c-1)}{2(d-1)}$.
\end{proof}
\end{Theorem}

\newpage
\section{Additional Analysis}
\subsection{Shrinking effect for unidimensional data}
\paragraph{Bounded decision region.}
Without loss of generality, let the decision region be interval $\db = [-R, R]$. By the symmetric nature of Gaussian smoothing, we see that $\db_\sigma$ is also an interval of the form $[-a, a]$. We claim that for large $\sigma$, $a < R$ and for even larger $\sigma$, $\db_\sigma$ disappears. Formally, we do the analysis as follows.

For the shrinking, we check the value of $f_\sigma(R)_1$. By definition \ref{s_def:gaussiansmoothing}, we see that $f_\sigma(R)_1 = \Phi(\frac{2R}{\sigma}) - \Phi(0)$ and if 
$$ \sigma > \frac{2R}{\Phi^{-1}(\frac{1}{2} + \frac{1}{c})},$$
$f_\sigma(R) < \frac{1}{c}$ is true. Thus, the bounded decision region of unidimensional data shrinks with smoothing factor $\sigma >  \frac{2R}{\Phi^{-1}(\frac{1}{c} + \frac{1}{2})}$. 

For the vanishing rate, we check the value of $f_\sigma(x)_1$ at $x= 0$. Now since $f_\sigma(0)_1 = \Phi(\frac{R}{\sigma}) - \Phi(-\frac{R}{\sigma})$, we have that if 
$$\sigma > \frac{R}{\Phi^{-1}(\frac{1}{2} + \frac{1}{2c})},$$
$f_\sigma(0)_1 < \frac{1}{c}$ is true, \textit{i.e.}, $\db_\sigma$ vanishes.

\paragraph{Semi-bounded decision region.}
In a  unidimensional case, our definition of semi-bounded regions degenerates into an interval $I$ of the form $[a,\infty)$.  In this case, Theorem~\ref{s_thm:semi3} gives a trivial bound of $0$ for the shrinkage of the decision region, suggesting that no shrinking happens. However, we emphasize that in practice, shrinking might still happen and more detailed analysis is left for future work.

\subsection{Bounded decision region behaviors}
\begin{figure}[h]
    \centering
    \includegraphics[width=0.5\linewidth]{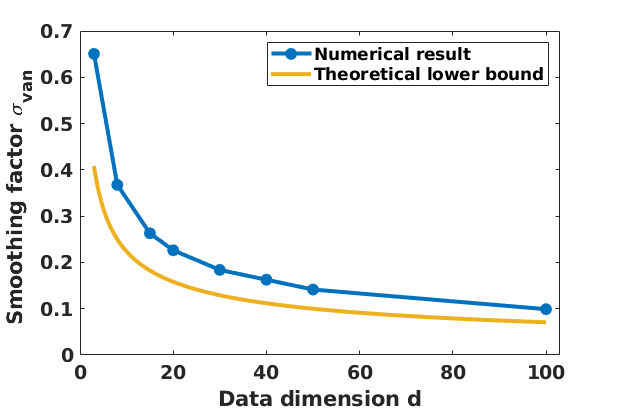}
    \caption{The vanishing smoothing factor $\sigma_{\text{van}}$ with an increasing input-space dimension in the exemplary adversarial ball.}
    \label{fig:s_vanish_dimension}
\end{figure}
\begin{figure}[h]
    \centering
    \includegraphics[width=0.45\linewidth]{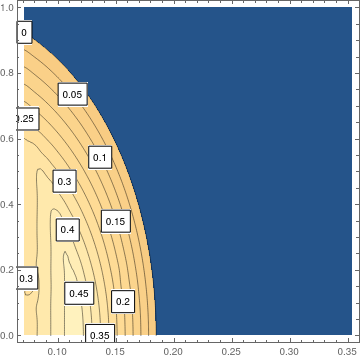}
    \caption{The certified radius of smoothed classifiers with an increasing input-space dimension when $d=30$.}
    \label{fig:s_cert_rad30}
\centering
    \includegraphics[width=0.55\linewidth]{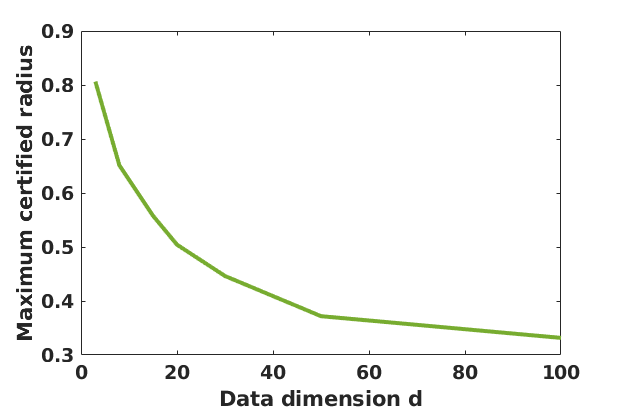}
    \caption{The maximum certified radius with an increasing input-space dimension in the exemplary case.}
    \label{fig:s_radius_dimension}
\end{figure}
The vanishing smoothing factors $\sigma_{\text{van}}$ with different data dimensions implied by Figure~2 of the main text together with the theoretical lower bound found in Theorem~\ref{s_thm:2} is given as Figure~\ref{fig:s_vanish_dimension}.

Figure~\ref{fig:s_cert_rad30} shows the certified radius behavior as a function of the distance of points from the origin (y-axis) and the smoothing factor $\sigma$ (x-axis) for dimension $d=30$. The contour lines in Figure~\ref{fig:s_cert_rad30} mark the certified radius of points under Gaussian smoothing. It is notable that points closer to the origin generally have larger certified radii and the certified radius of the point at the origin (y-axis $y=0$) drops to zero at vanishing smoothing factor $\sigma_{\text{van}}=0.184$ as specified in Figure~\ref{fig:s_vanish_dimension}. Specifically, one can readily verify that the certified radii of points closer to the origin increase with the growing smoothing factor $\sigma$ but begin to decrease at certain point, which is coherent with our observations through Figure~3(a) of the main text. Conducting similar experiments for different dimensions completes the maximum certified radius vs. data dimension relationship as shown in Figure~\ref{fig:s_radius_dimension}.

\subsection{Semi-bounded decision region certified radius behaviors w.r.t data dimensions}
\begin{figure}[t]
    \centering
    \includegraphics[width=0.5\linewidth]{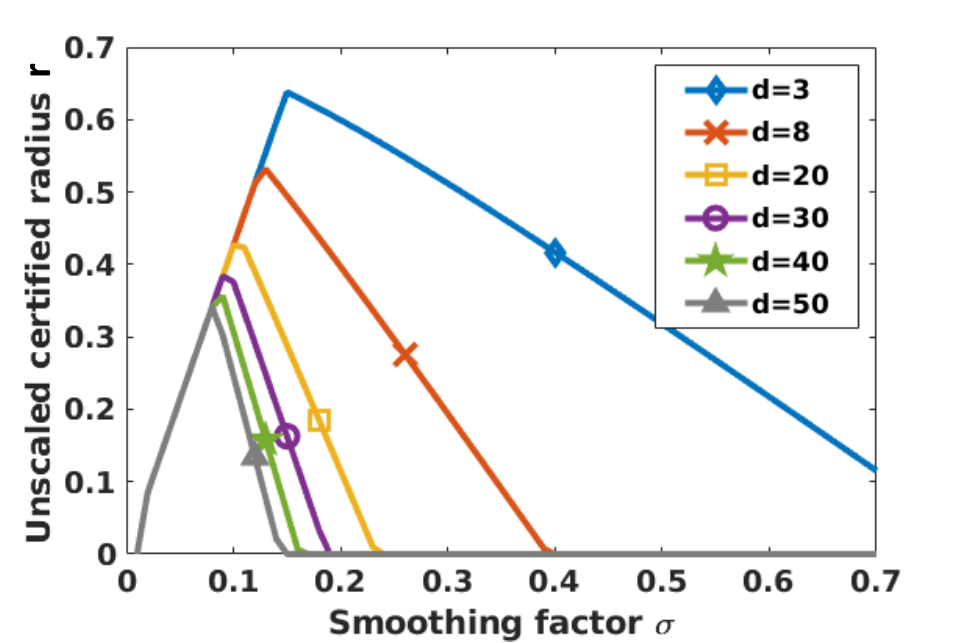}
    \caption{The unscaled certified radius $r$ of a point on the axis $v$ for different input data dimension $d$.}
    \label{fig:s_semi_radius_dimension}
\end{figure}
In Figure~\ref{fig:s_semi_radius_dimension}, we show the unscaled certified radius $r$ as a function of an increasing smoothing factor $\sigma$ for different input data dimension $d$ with fixed narrowness $\theta=45^\circ$. One can then see similar trend as told in Figure~3(a) of the main text in the bounded decision region case, the maximum certified radius (the peak) also decreases with the increasing dimension.


\end{document}


%

%

\onecolumn
\aistatstitle{Instructions for Paper Submissions to AISTATS 2021: \\
Supplementary Materials}

\section{FORMATTING INSTRUCTIONS}

To prepare a supplementary pdf file, we ask the authors to use \texttt{aistats2021.sty} as a style file and to follow the same formatting instructions as in the main paper.
The only difference is that the supplementary material must be in a \emph{single-column} format.
You can use \texttt{supplement.tex} in our starter pack as a starting point, or append the supplementary content to the main paper and split the final PDF into two separate files.

Note that reviewers are under no obligation to examine your supplementary material.

\section{MISSING PROOFS}

The supplementary materials may contain detailed proofs of the results that are missing in the main paper.

\subsection{Proof of Lemma 3}

\textit{In this section, we present the detailed proof of Lemma 3 and then [ ... ]}

\section{ADDITIONAL EXPERIMENTS}

If you have additional experimental results, you may include them in the supplementary materials.

\subsection{The Effect of Regularization Parameter}

\textit{Our algorithm depends on the regularization parameter $\lambda$. Figure 1 below illustrates the effect of this parameter on the performance of our algorithm. As we can see, [ ... ]}

\vfill